\documentclass[preprint,12pt]{elsarticle}

\usepackage{amssymb}

\usepackage[latin1]{inputenc}   
\usepackage[T1]{fontenc}
\usepackage{amsmath}
\usepackage{amssymb,amsmath}
\usepackage{graphicx}
\usepackage{graphics}
\usepackage{multirow}
\usepackage{color}
\usepackage{layout}  				
\usepackage{pdfpages} 				
\usepackage{pdflscape}
\usepackage{multirow}
\usepackage{tikz}
\usepackage{array}%
\usepackage[linesnumbered]{algorithm2e}
\usepackage{graphics} 
\usepackage{epsfig} 
\usepackage{amssymb,amsmath} 
\usepackage{algpseudocode} 
\usepackage{color}
\usepackage[numbers]{natbib}
\usepackage{lscape}
\biboptions{authoryear}

\journal{ESWA (	https://doi.org/10.1016/j.eswa.2021.115638)}

\begin{document}
	
	\begin{frontmatter}
		
		\title{An Efficient Multi-objective Evolutionary Approach for Solving the Operation of Multi-Reservoir System Scheduling in Hydro-Power Plants}
		
		\author[my1b,my1] {C. G. Marcelino \corref{mycorrespondingauthor}}
		\cortext[mycorrespondingauthor]{Corresponding author}
		\ead{carolina@ic.ufrj.br}
		
		\author[my1,my2]{G.M.C. Leite}\ead{gmatos@cos.ufrj.br}
		
		\author[my1]{C.A.D.M. Delgado}\ead{carla@ic.ufrj.br}
		
		\author[my1c]{L.B. de Oliveira}\ead{lucas.braga.deo@gmail.com }
		
		\author[my1c]{E.F. Wanner}\ead{efwanner@cefetmg.br}
		
		\author[my1b]{S. Jim\'enez-Fern\'andez}\ead{silvia.jimenez@uah.es} 
		\author[my1b]{S. Salcedo-Sanz}\ead{sancho.salcedo@uah.es}  
		
		\address[my1b]{Department of Signal Processing and Communications, Universidad de Alcal\'a, Spain.}
		
		\address[my1]{Institute of Computing, Federal University of Rio de Janeiro, Brazil.}
		
		\address[my2]{Post-Graduate Program in Systems Engineering and Computer Science, Federal University of Rio de Janeiro, Brazil.}
		
		\address[my1c]{Computing Department, Federal Center of Technology Education of Minas Gerais, Brazil. }


		\begin{abstract}

			This paper tackles the short-term hydro-power unit commitment problem in a multi-reservoir system - a cascade-based operation scenario. For this, we propose a new mathematical modelling in which the goal is to maximize the total energy production of the hydro-power plant in a sub-daily operation, and, simultaneously, to maximize the total water content (volume) of reservoirs. For solving the problem, we discuss the Multi-objective Evolutionary Swarm Hybridization (MESH) algorithm, a recently proposed multi-objective swarm intelligence-based optimization method which has obtained very competitive results when compared to existing evolutionary algorithms in specific applications. The MESH approach has been applied to find the optimal water discharge and the power produced at the maximum reservoir volume for all possible combinations of turbines in a hydro-power plant. The performance of MESH has been compared with that of well-known evolutionary approaches such as NSGA-II, NSGA-III, SPEA2, and MOEA/D in a realistic problem considering data from a hydro-power energy system with two cascaded hydro-power plants in Brazil. Results indicate that MESH  showed a superior performance than alternative multi-objective approaches in terms of efficiency and accuracy, providing a profit of \$412,500 per month in a projection analysis carried out.
			
		\end{abstract}
		
		
	\end{frontmatter}
\section{Introduction} \label{sec:introduction}
Hydro-power is one of the most important sustainable energy sources in countries with a huge fluvial resource, such as Brazil. The water resources management, combined with the growth in demand for electricity and climate change are impacting factors in the flow regime of rivers, directly interfering in the development of economic activities for the production of hydro-electric energy. Compared to other renewable resources, hydro-power has exceptional advantages, such as the ability to generate electricity without producing any pollution and to provide water flow control in the rivers \citep{sharma}. A big challenge in hydro-power is the modelling and operation of systems that generates energy using two or more hydro-power plants (HPPs) in a cascade process. This method of conducting the electric dispatch production is known as the Operation of Multi-Reservoir System (OMRS) \citep{roefs}. As in the case of single hydro-power plants, in OMRS it is needed to define an optimal schedule for the production units, usually on an hourly basis, to maximize the electrical power obtained from a given water volume. In the case of multi-reservoirs, the optimization problem is usually very hard with non-linear objective functions, an extremely large search space dimension and, in many occasions, several objectives with different constraints to be fulfilled \citep{barros}. 

There are different methods to solve OMRS problems described in the literature, that can be roughly divided into two classes: conventional methods and bio-inspired meta-heuristics. In general, conventional methods are to some extent deterministic algorithms. \cite{cai} and \cite{Yoo} used Linear Programming (LP) methods to maximize hydro-power generation. The works by \cite{zeng} and \cite{catalao} addressed the maximization of reservoir volume with use of Nonlinear Programming. Mixed-integer Linear Programming (MILP) is another classic tool used to minimize the maintenance costs and usage of water in hydro-power plants in \cite{canto},\cite{ge} and \cite{chen}. Methods based on Lagrangian Relaxation to minimize the total costs of production were discussed by \cite{guan} and \cite{finardi}. Dynamic Programming techniques were also adopted to obtain the optimal management operation as proposed in \cite{rizzo}. Fuzzy models have also been applied to conduct the dispatch operation in HPPs, as in \cite{moe} and \cite{zhangg}.

Bio-inspired algorithms have also been successfully applied to different problems in hydro-power. For example, \cite{naresh} and \cite{xie} used different types of neural networks to solve hydro-scheduling, a subproblem of OMRS.  Genetic Algorithms (GAs) have been used to solve several types the electric dispatch in OMRS: to provide optimal operation of these type of facilities \citep{leite}, to maximize the power production for a case scenario in Turkey \citep{cinar} and to maximize the power production of small communities in Honduras \citep{tapia}. Solutions inspired on swarm intelligence adopting the Particle Swarm Optimization (PSO) have been applied to minimize the use of water in power generation as described on \cite{lu}. PSO has been also used to minimize environmental impacts of power generation \citep{xinli} and to minimize the production costs \citep{mandal}. Considering the ecological environment problem described in \cite{fanzou}, the authors have applied a Differential Evolution (DE) algorithm to solve the electrical dispatching problem. DE algorithm versions were also applied to maximize the volume of water in reservoirs \citep{guedes}. 

Related works have presented high degrees of success in these  practical engineering problems. However, there are still certain weaknesses when conventional or bio-inspired techniques are used to solve OMRS related problems. In many cases, LP methods failed to address the widespread nonlinearity in the basic feature information of HPP reservoirs \citep{cai}. Nonlinear programming often showed inaccuracies due to linearization of nonlinear constraints when addressing the non-convex objective function in HPP system \citep{zeng}. Sometimes applying the dynamic programming approaches can be mainly limited by dimensionality problem in which OMRS is involved \citep{rizzo}. The great challenge for neural network methods is conducting the selection of computational parameters \citep{naresh}, a time consuming task mostly incompatible with the real time nature of the dispatch problem. Population-based and bio-inspired methods based on evolutionary algorithms can be easily trapped in local optima due to the premature convergence problem \citep{za,guedes}. 

In addition to these issues related to the algorithms focused on the OMRS problem, note that the short-term HPP scheduling problem in OMRS can have more than one objective: some plant operators might need not only to maximize the efficiency of the energy production process, but also to keep the turbine flow close to a target value or to optimize the water balance between the reservoirs. Despite the multi-objective nature of the problem, the majority of existing methods perform a scalarization to transform the problem into a single-objective one. Nevertheless, solving a multi-objective problem via a scalarized mono-objective approach can lead to a crucial information loss \citep{jors}. 

A wide variety of multi-objective Evolutionary Algorithms (MOEAs) have been proposed and successfully applied to many real-world optimization problems \citep{zhou2011multiobjective}. Some MOEAs employ the concept of Pareto Dominance to find a set of non-dominated solutions, which represent a set of efficient solutions considering the objective functions of the problem at hand. As an example, we can cite the Non-dominated Sorting Genetic Algorithm (NSGA-II) \citep{nsga2}, the Strength Pareto Evolutionary Algorithm (SPEA2) \citep{spea2} and the Multi-objective Particle Swarm Optimization (MOPSO) \citep{mostaghim}, to name a few. Despite the popularity in academia, the use of MOEAs in industry is not so common. Specifically for the OMRS scenario, very few works tackle the problem in its multi-objective formulation. The NSGA-II has been applied to maximize the river habitat quality and hydro-power generation \citep{gallerano}. 

The SPEA2 has been used in \cite{hidalgo} to minimize daily release from the plant and the number of times that the status of the unit generator is changed. An approach using the Improved Partheno Genetic Algorithm (IPGA) has been conducted to optimize a system with two HPPs in China \citep{jwang}. The use of PSO to solve complex multi-dimensional problems has grown significantly due to its simplicity and easy applicability. Currently, new algorithms inspired by swarm intelligence have been widely adopted for solving highly nonlinear, multi-modal, NP-Hard and multi-objective problems, and have proven successful in those cases \citep{alm}. Some works that follow these approaches are: the Improved Multi-Objective Particle Swarm optimization (IMOPSO) algorithm proposed by \cite{hu}, a Multi-objective Particle Swarm Optimization (MOPSO) version \citep{feng} and a Parallel Multi-PSO (PMPSO) described in \cite{niu}. Specifically, swarm-based algorithms proved to be very efficient and fast for solving problems in the energy field \citep{pes17, cec18, viab}. Thus, in this work we propose a hybrid swarm algorithm, aiming to use the best mechanisms coming from evolutionary computation within the well-founded framework inherent to swarm intelligence.

In the last years, other Pareto dominance-based MOEAs have been proposed to deal with problems having three or more objectives. Recently, a large number of specialized algorithms have been proposed and applied to different topics such as
big data optimization \citep{nr6}, cyber-physical social systems \citep{nr1}, interval multi-objective optimization problems \citep{nr2}, distributed manufacturing problems \citep{nr5}, vehicle routing \citep{nr3}, signal processing \citep{nr4}, and correlated subjects \citep{nr7}. 
In this paper we will further discuss the performance of two MOEAs that have proven to be powerful to deal with problems with any number of objectives: the MOEA based on decomposition (MOEA/D, by \cite{nr8}) and  the reference-point based non-dominated sorting algorithm (NSGA-III \citep{nr9,nr10}). These are standard, baseline algorithms, on top of which further approaches have been proposed.

MOEA/D \citep{nr8} is a
decomposition-based MOEA that emphasizes convergence and
diversity of population. The problem  is
decomposed into a set of subproblems and then optimized
simultaneously. A uniformly generated set of weight vectors associated with a fitness assignment method is usually used to decompose the original problem. Improved and blended versions of MOEA/D have been proposed in the literature. An improvement proposed in \cite{nr11} using the Information Feedback Models (IFM) demonstrated competitive results when compared to the standard version in a set of large scale benchmark functions. A modified multi-objective evolutionary algorithm with decomposition plus random local search (MMOEA/D-RL) was proposed in \cite{nr5} to solve a distributed manufacturing problem. The central idea of MMOEA/D-RL is that the weight vectors are initialized randomly, and then the neighbors of each solution are determined accordingly. Sophisticated procedures are used to improve the algorithm performance and the results have showed a competitive performance when applied to a real-world problem.

Another decomposition-based algorithm using a localized control variable analysis approach (called LSMOEA/D) was proposed by \cite{nr12}. In the LSMOEA/D method, the guidance of reference vectors is incorporated into the control decision variable analysis, leading  to a competitive performance when solving benchmark problems. 
\cite{nr13} proposed an algorithm to improve the speed of convergence in large-scale multi-objective problems. In benchmark functions with 5000 decision variables, the large-scale evolutionary multi-objective algorithm assisted by directed sampling (LMOEA-DS) showed competitive results when solving problems with three conflicting objectives. However the authors concluded that the LMOEA-DS suffers from a common weakness of decomposition-based algorithms: their performance heavily depends on the degree of match between the distribution of the reference solution and the offspring. 

NSGA-III algorithm is a domination-based MOEAs in which the domination principle plays a key role. In its famous counterpart, the NSGA-II, the  crowding distances of all individuals are calculated at each generation and used to maintain the population diversity. Inheriting the non-dominated sorting from NSGA-II, in the NSGA-III the reference points are employed to keep the diversity. The NSGA-III  has been used to solve various type problems such as information feedback models \citep{nr14} and large-scale optimization problems \citep{nr15}. Improved and blended approaches have also been proposed in the literature to solve different problems. An improvement using the Information Feedback Models (IFM) scheme obtained competitive results in solving large-scale many-objective problems \citep{nr14}. In the same way, refinements in the NSGA-III can be seen with use of simulated binary, uniform, and single point type crossover \citep{nr15}.

Development of algorithms deriving from MOEA/D and NSGA-III has gain attention in the last years. From the best of our knowledge, those algorithms have not yet been applied to solve the OMRS problem. Motivated by this fact and by the successful real-world applications, standard versions of MOEA/D and NSGA-III have been used to assess the performance of our proposed approach. Moreover, since the OMRS problem studied in this paper has a bi-objective nature, the well-known and successful MOEAs, NSGA-II and SPEA-2, have also been included in the performance assessment carried out.

Although the maximization of energy production in OMRS can be modelled by a common objective function, the maximization of the volume of simultaneous reservoirs is yet unexplored. In this work, we deal with these two conflicting objectives, aiming to guarantee the maximum efficiency of each turbine-generator set, while taking into account the hydraulic losses of the system. Keeping this in mind, this work proposes a new MOEA applied to the short-term dispatch of an HPP in OMRS. Our analysis takes into account a cascading system composed of two hydro-power reservoirs serving multiple interconnected power plants in Brazil. 

To solve the electric dispatch in OMRS operation process, the novel Multi-objective Evolutionary Swarm Hybridization (MESH) is proposed and discussed. MESH is based on C-DEEPSO \citep{carolapp,jors}, a mono-objective evolutionary algorithm with recombination rules borrowed from PSO or, alternatively, a mono-objective swarm optimization method with selection and self-adaptive properties. The rationale here is due to the performance superiority of C-DEEPSO when applied to mono-objective versions of diverse power systems problems. Taking advantage of swarm intelligence methods and coupled with operators from evolutionary computation techniques, the proposed approach is compared with four algorithms, NSGA-II, SPEA2, NSGA-III, and MOEA/D. The experimental results show that MESH is extremely competitive in solving the short-term electric dispatch to HPPs in the multi-reservoir operation system. Therefore, MESH acts as an electrical dispatch controller system capable of offering optimized solutions for the daily planning horizon. Furthermore, MESH guarantees the maximal production with a good use of water resources, since the obtained solutions  are able to maximize the water volume of the reservoirs. This characteristic differentiates MESH from the other techniques previously discussed.
More specifically, this paper presents the following contributions: 
\begin{itemize}
	\item a novel mathematical modelling for the hydro-power unit commitment in a multi-reservoir system finding optimal water discharge and power;
	\item a Multi-Objective Evolutionary Swarm Hybridization (MESH) algorithm to solve the proposed unit commitment problem;
	\item the proposed approach, MESH, has been compared to a set of benchmark problems and the results indicate a competitive performance; 
	\item usage of a realistic data from a Brazilian hydro-power energy system with two HPPs in a cascade scenario;
	\item an in-depth performance assessment of MESH comparing to four different and well-known algorithms, NSGA-II, SPEA-2, MOEA/D, and NSGA-III in the hydro-power unit commitment problem;
	\item obtained results indicate a competitive performance favoring MESH in terms of efficiency and accuracy when applied to the hydro-power unit commitment problem;
	\item a projection analysis has been carried out indicating a profit of \$412,500 per month solving the problem using the proposed approach.
\end{itemize}

The rest of the paper has been organized as follows: Section \ref{meshalg} describes the mechanisms of the MESH as an hybrid method able to solve continuous problems as the electric dispatch in the OMRS operation process. Section \ref{model} details the short-term electrical dispatch mathematical modelling of hydro-power plants in cascade operation. Section \ref{exper} comprises the experiment of MESH using continuous benchmark functions, and the comparative analysis of MESH with other methods. Section \ref{realp} shows the experiments results of short-term multi-objective electric dispatch in cascade operation. Finally, Section \ref{finalr} illustrates the final remarks regarding the overall MESH performance.

\section{Multi-objective Evolutionary Swarm Hybridization algorithm}\label{meshalg}

Evolutionary algorithms (EAs), a popular class of meta-heuristics in the area of optimization research, are techniques inspired by the processes of biological evolutionary structures. In a multi-objective view, EAs are able to provide feasible solutions for two or more objectives at the same time. Currently, new approaches are being developed in a merged way, which can also be considered as methodologies to hybridize. These hybrid methods consider mixing better operators of different algorithms to obtain a more efficient optimization tool. In this context, the combination of Differential Evolution (DE), Particle Swarm Optimization (PSO), and the sorting operator from the Non-dominated Sorting Genetic Algorithm (NSGA-II) represents a promising way to create superior optimizers in multi-objective optimization problems. 

\begin{figure}[!ht]
	\begin{center}
		\includegraphics[scale=0.5]{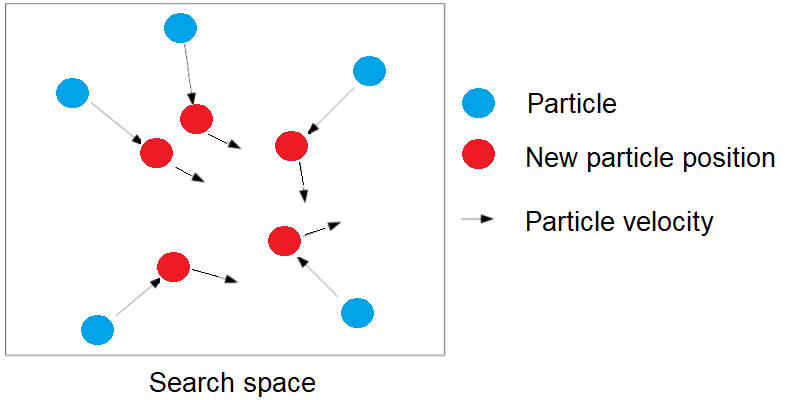}
		\caption{Swarm with the respective positions and velocities of a given iteration, in a two-dimensional search space. \label{si}}
	\end{center}
\end{figure}

Motivated by the competitive performance of the previously proposed C-DEEPSO algorithm \citep{carolapp,jors} in different problems related to power systems solutions, in this work we propose a novel hybrid algorithm for multi-objective problems, the  Multi-objective Evolutionary Swarm Hybridizion (MESH). In swarm optimization, the exploration of the search space made by a particle aims to follow the best solutions already found both in the particle itself and in its neighborhood, allowing to scan the search space and find new solutions for better evaluation. The exploration is carried out by updating the positions and velocities of the particles at each iteration (see Figure \ref{si}). The process is repeated for a pre-defined number of iterations or until a pre-defined convergence criterion is reached.

The success of the search for an optimal position of a particle depends not only on the performance of the particle individually, but also on the information shared with the swarm. This joint skill of the swarm has been attributed to the concept of Swarm Intelligence. The swarm optimization to solve complex multidimensional problems has grown significantly due to its simplicity and easy applicability. 

\begin{figure}[!ht]
	\begin{center}
		\includegraphics[scale=1]{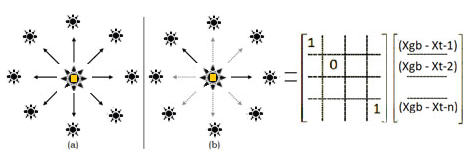}
		\caption{C-DEEPSO topology: Star communication (a) and Stochastic
			star topology (b). The binary matrix $\mathbf{C}$ is obtained using the
			rule: randomly generate $N$ values within the $[0,1]$ interval for
			each dimension inside of each solution. The randomly generated
			value is compared to communication rate $P$. If this random
			value is greater than $P$, the element $C_{ij}$ of $\mathbf{C}$ matrix
			receives 0, otherwise 1.\label{starf}}
	\end{center}
\end{figure}

In this context, the MESH method proposed here has been initially developed to represent problems in a continuous search space. In MESH, the recombination is governed by the so called Movement Rule, in the same way as in the C-DEEPSO algorithm. This rule is given by Equations (\ref{equ1}) and (\ref{equ2}):

\begin{eqnarray}\label{equ1}
	\mathbf{V}_{n} = \mathbf{w}_{I}^{*} \mathbf{V}_{n-1} + \mathbf{w}_{A}^{*} (\mathbf{X}_{sn} - \mathbf{X}_{n-1}) + \mathbf{w}_{C}^{*} \mathbf{C} (\mathbf{X}_{gb}^{*} - \mathbf{X}_{n-1}), 
\end{eqnarray}
\begin{eqnarray}\label{equ2}
	\mathbf{X}_{n} = \mathbf{X}_{n-1} + \mathbf{V}_{n}, 
\end{eqnarray}

in which $\mathbf{X}_{sn}$ is a position obtained by using the recombination mechanisms of Differential Evolution (DE). The subscript $n$ denotes the current generation. $\mathbf{X}_{n}$ is the current particle or solution. The term $\mathbf{X}_{gb}$ addresses the best solution ever found by the population. $\mathbf{V}_{n}$ is the velocity of the individual. The term $\mathbf{C}$ represents a $N \times N$ diagonal matrix of random variables sampled in every iteration, according to a Bernoulli distribution with success probability $P$, as described in Figure \ref{starf} that exemplifies the ``star topology'' proposed. MESH has a memory archive file (the $MB$) in which a subset of the best solutions from the last population is stored. The superscript $^*$ indicates the corresponding parameter that undergoes evolution under a mutation process. Typically, the mutation of a generic weight $w$ of an individual follows a simple additive rule as described by Equation (\ref{equpesos}),
\begin{eqnarray}\label{equpesos}
	\mathbf{w}^{*} = \mathbf{w} + \tau \times \mathcal{N}(0,1), 
\end{eqnarray}
in which $\tau$ is the mutation rate that must be set by the user. $\mathcal{N}(0,1)$ is a number sampled from the standard Gaussian Distribution. Mutation of $\mathbf{X}_{gb}$, which is carried out for every particle, is performed according to:
\begin{eqnarray}\label{equgb}
	\mathbf{X}_{gb}^{*} = \mathbf{X}_{gb}[1 + \tau \times \mathcal{N}(0,1)]. 
\end{eqnarray}

MESH uses the Movement Rule from C-DEEPSO with a multi-objective approach for handling two goals. Basically, in memory ($MB$) the MESH algorithm employs the non-dominated sorting operator from NSGA-II to identify and update the Pareto frontier throughout the search process. The solutions in this memory are used in turn as the new attractors $\mathbf{X}_{sn}$ from Equation (\ref{equ2}). The memory is updated on each iteration by combining the Pareto front of the population with the non-dominated solutions stored. The sorting operator of NSGA-II is applied to this augmented set of solutions aiming to identify the non-dominated ones. If the Pareto front is larger than the maximum memory size, then the crowded-distance from the NSGA-II operator is applied to keep the memory size.  

Inspired by the guide particle concepts from \citep{mostaghim}, the MESH has a process to obtain guides according to different solutions: Individual Guide ($G_{i}$), that is the set of the best solution found by the particle (the choice of which particle from individual guide array to use is randomly selected) and Swarm Guide ($G_{s}$), that corresponds to a solution found by the swarm that is greater than the current particle solution. The swarm guide is applied in the memory archive or in the current swarm. The $G_{s}$ is calculated by using Equation (\ref{equ3}):
\begin{eqnarray}\label{equ3}
	\boldsymbol{\sigma} = \begin{bmatrix}
		(f_{1}^{2} - f_{2}^{2})/(f_{1}^{2} + f_{2}^{2} + f_{3}^{2}) \\ 
		(f_{2}^{2} - f_{3}^{2})/(f_{1}^{2} + f_{2}^{2} + f_{3}^{2})\\ 
		(f_{3}^{2} - f_{1}^{2})/(f_{1}^{2} + f_{2}^{2} + f_{3}^{2})
	\end{bmatrix}.
\end{eqnarray}

\begin{figure}[!ht]
	\begin{center}
		\includegraphics[scale=0.5]{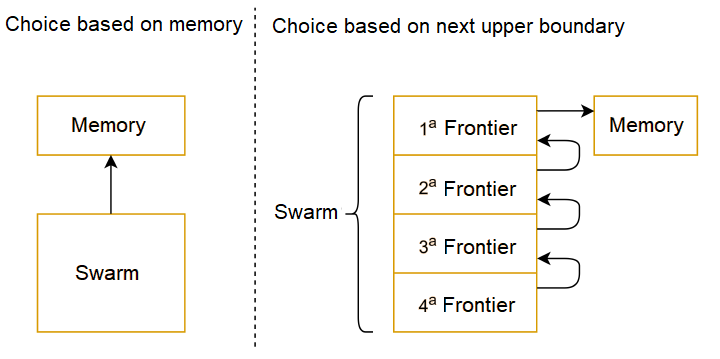}
		\caption{Diagram for choosing a particle swarm guide in MESH. In the method shown on the left, all particles of the population choose from memory. In the method shown at right, the choice is made based on the next upper boundary in relation to that this particle belongs. The first boundary particles in turn use memory.\label{f1}}
	\end{center}
\end{figure}

Equation (\ref{equ3}) refers to the Sigma method, proposed in \citep{mostaghim}. The $\boldsymbol{\sigma}$ assigns a value to each particle in the swarm to estimate the distance in the objective functions space.  All solutions belonging to the search space that are in same line will receive the same sigma ($\boldsymbol{\sigma}$) value. The idea behind the $\boldsymbol{\sigma}$ method is to use the particle suitability values in each objective function as the coordinates, thus the global best for a particle is another particle with the nearest sigma coordinates. Therefore,  Equation (\ref{equ3}) exemplifies how the sigma coordinates are calculated for a three dimensional objective space. Specifically, in MESH, another alternative is to combine the sigma method with the best overall choice procedure. In this process, the particle swarm guide is the one that is closest to the next upper boundary of the current one. If the current boundary of the particle is the first one, the choice will be made with memory ($MB$). It is shown in Figure \ref{f1}.

\begin{figure*}[!ht]
	\begin{center}
		\includegraphics[scale=0.4]{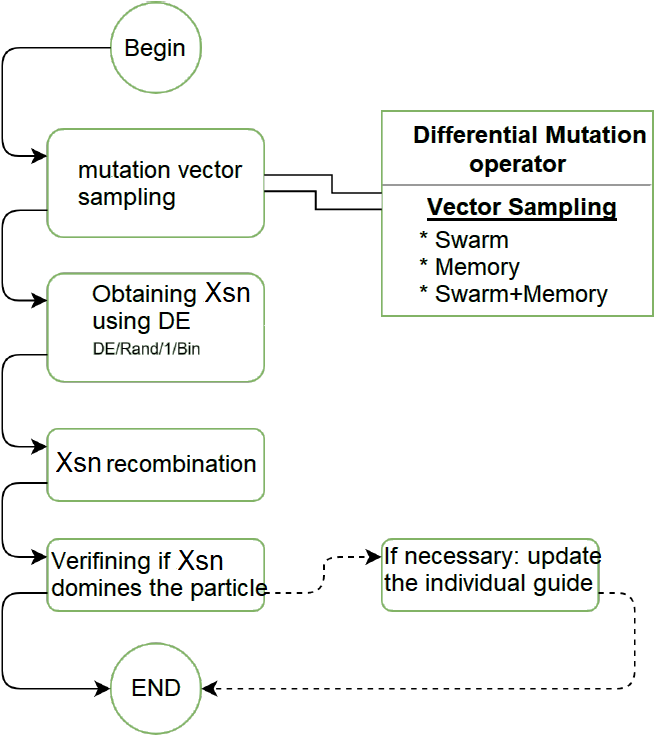}
		\caption{Diagram of operation of the MESH differential mutation operator. The sampling can occur in three forms: from the swarm (current population), using solutions from memory ($MB$) or in a mixed approach using solutions from the swarm and memory. \label{mutationfig}}
	\end{center}
\end{figure*}

The vectors used for differential mutation in MESH are sampled from three different groups: a group containing only particles from the front population equal to or greater than the particle, a group with memory particles, or a combination of the previous two groups. In this work, the DE/Rand/1/Bin strategy is implemented. The diagram of the differential mutation operator is shown in Figure \ref{mutationfig}.
General functioning of MESH is described in Algorithm (\ref{meshp}). In this work, we adopted the non-dominated sorting procedure and the crowd-distance operator proposed in \citep{nsga2}. Algorithm (\ref{individualguide}) shows the pseudo-code to update individual guide array. 


\begin{algorithm}[!ht]
	
	{\scriptsize
		\DontPrintSemicolon
		\Begin{
			\textbf{Set} values for parameters of MESH -- Population size $NP$, Mutation rate $\tau$, Communication rate $P$, Memory size $MB$, Mutation Type, and Guide type, Individual Guide size;\; 
			\textbf{Evaluate} the current population, $NP$;\;
			\textbf{Update} individual guide with current solution;\;
			\textbf{Apply} dominance mechanism in the population;\; 
			\textbf{Update} the Memory ($MB$):  merge memory with the first frontier of the non-dominated set;\; 
			\textbf{Apply} dominance mechanism in the Memory ($MB$) and \textbf{Apply} crowd-distance operator if the frontier is bigger than memory size;\; 
			\nl\While{stopping criteria is not satisfied}{
				{\textbf{Apply} mutation mechanism into swarm};\;
				\If{the mechanism needs a swarm guide}
				{ 
					{ \textbf{Update} swarm guide using Equation (\ref{equ3})};\;
				}
				\If{ any $\mathbf{X}_{sn}$ dominates its current individual update }
				{ 
					{ $\mathbf{X}_{sn}$  replaces the current individual};\;
					{Apply dominance mechanism};\;
					{Update the memory $MB$};\;
				}
				{\textbf{Update} swarm guide using Equation (\ref{equ3})};\;
				{\textbf{Copy} the current swarm};\;
				{\textbf{Mutate} the strategy parameters in swarm and copy $\mathbf{w}_{I}$, $\mathbf{w}_{A}$, $\mathbf{w}_{C}$ using Equation (\ref{equpesos})};\;
				{\textbf{Mutate} $\mathbf{X}_{gb}^{*}$ using Equation (\ref{equgb}) in current swarm and copy};\;
				{\textbf{Apply} movement rule in current swarm and copy using Equation (\ref{equ1})};\;
				{\textbf{Update} individual guide using Algorithm (\ref{individualguide})};\;
				{\textbf{Apply} dominance mechanism in the swarm and copy};\;
				{\textbf{Select} the best particles based on frontiers and \textbf{if} necessary \textbf{Apply} crowd-distance };\;	
				{\textbf{Update} the Memory $MB$};\;
				{T=T+1};\;
			}
	}}
	\caption{{\it Pseudo-code of MESH.} \label{meshp}}
	
\end{algorithm}

\begin{algorithm}[!ht]
	
	{\scriptsize
		\DontPrintSemicolon
		\Begin{
			\eIf{Current particle dominates the particles in individual guide array}
			{ 
				{All particles in individual guide array are removed and the current particle is added};\;
			}
			{
				\If{Current particle is not dominated by the particles in individual guide array and it also don't dominate those particles}
				{ 
					{Current particle is added to the individual guide array};\;
				} 
			} 
	}} 
	\caption{{\it Pseudo-code to update individual guide array} \label{individualguide}}
	
\end{algorithm}	

According to  \cite{kra}, it is now well established that pure population-based algorithms are not well suited to refinement of complex spaces and that hybridization with other techniques can significantly improve search efficiency. In this way, the MESH algorithm joints different concepts of swarm intelligence and evolutionary optimization to be a viable approach to solve real world problems. The MESH algorithm allows an efficient combination of the PSO and DE algorithms, as it employs typical inspiration/recombination at the swarm intelligence inherited from PSO and the mutation rules present in DE. In addition, it incorporates the dominance ordering process (dominance mechanism) and crow-distance operation observed in the NSGA-II (see \citep{nsga2}), as well as the use of the swarm guides process, to escape from non-promising regions in the search space. 

A computational MESH complexity analysis (see Algorithm \ref{meshp}) has been carried out.  Let  the  population size, number of objectives, number of decision variables (dimension), and memory size be $NP$, $M$, $D$ and $MB$,  respectively. The population is sorted by dominance (Steps $(5)$, $(14)-(18)$ and $(25)$) with time complexity $O(M\times N\times \log(N))$. In Step $(10)$, mutation mechanism is executed with time complexity $O(D\times NP\times MB)$. Each individual guide update (Step $(24)$) has time complexity $O(NP\times D)$. Mutations (Steps $(21)$ and $(22)$) are executed with time complexity $O(D^2)$. Movement rule (Step $(23)$) is applied to each particle with constant time, leading to a time complexity of $O(NP)$. Finally, the next generation is selected with time complexity $O(M\times NP^2 \times \log(NP))$ and memory is updated with time complexity $O(M\times NP \times \log(NP))$. From the above results, after omitting the low-order terms, total time complexity of MESH algorithm is $O(T\times M\times NP^2 \times \log(NP))$, which is polynomial in $NP$. A complete MESH code version is available at https://github.com/gabrielmatos26/MESH.

Therefore, addressing our proposing in a full view, MESH is a hybrid algorithm that incorporates the movement rule from PSO, mutation scheme from DE, and the non-dominated sorting mechanism from NSGA-II. The central idea is to use swarm intelligence coupled with operators from evolutionary computation. MESH includes a swarm guide mechanism, with two options to choose from: (1) it uses the information from the best space positions saved in a memory population (that keeps part of the best individuals at each generation) or (2) the best solution found on the non-dominated Pareto front. Its mutation operator contemplates sampling from both the current swarm and vectors saved in memory, with the option of selecting vectors from both populations in this process. To explore the space MESH makes use of the evolutionary strategies inherent in DE. Therefore, the combination of these mechanisms (swarm guide and mutation operation) makes it capable of carrying out a more specialized search in the attraction basin without keeping the population trapped. MESH is built to solve continuous problems. In this work we present preliminary results in benchmark functions and we adapt its operation to solve the electrical dispatch problem in the hard OMRS process in hydroelectric plants. 

\section{Hydro-power Dispatch Problem: a OMRS in cascade mode}\label{model}

The operation planning purpose of an electric power system is to meet the requirements of cost, reliability, and optimal consumption of energy resources. In hydroelectric systems, such as the Brazilian system, the correct use of energy, available in limited quantities in the form of water in the reservoirs, is a problem with a very complex characteristic. The compromise between immediate decisions and the future consequences of these decisions makes the problem challenging and highlights the importance of proper planning. In this work, the time horizon adopted is the daily schedule, which is a problem of local operation with the operators of the plants and is considered a short-term process.

The planning operation of cascade hydroelectric systems (using the OMRS approach) is a particularly challenging problem, due to the complexity of its modeling and its characteristic of spatial and temporal coupling. Decisions to operate in a reservoir directly affect the levels of the other reservoirs downstream, and decisions about the storage or use of water affect the future level of the reservoirs, which may lead to a risk of deficit or leakage. Therefore, the operation of a hydroelectric systems must focus, in addition to the electrical operation, on the issue of the operation of the reservoirs, which leads to a problem of space and time coupling, i.e. a dynamic problem. The electrical dispatch producing of hydro-power plants is a typical problem in the optimal fields of OMRS. Attaining optimal operation rules are crucial for making the most of the comprehensive benefits. Thus, this work proposes a new mathematical modeling to provide electric production in cascade mode based on the mathematical model described in  \cite{carolmult, energi}. Here we have improved the previous modelling, taking into account the reservoir parameters (see Table \ref{not} for the modelling notation). In the proposed electric dispatch model described in this work, the power production, in $MW/h$, is obtained by Equation (\ref{meq12}):
\begin{eqnarray}\label{meq12}
	ph_{uj,t} = \text{g}\times \text{k} \times [\rho0_{uj} + \rho1_{uj}hl_{uj,t} + \rho2_{uj}Qt_{uj,t} + \rho3_{uj}hl_{uj,t}Qt_{uj,t} 
\end{eqnarray}
\[+ \rho4_{uj}hl_{uj,t}^{2} + \rho5_{uj}Qt_{uj,t}^{2}] \times [Hb_{u,t} - \Delta_{Huj,t}] \times Qt_{uj,t},\]

\begin{eqnarray}\label{meq13}
	hl_{uj,t} = Hb_{uj,t} - \Delta_{Huj,t}, 
\end{eqnarray}

\begin{eqnarray}\label{meq14}
	Hb_{uj,t} = fcm_{u,t} - fcj_{u,t},
\end{eqnarray}

\begin{eqnarray}\label{meq50}
	fcm_{u,t} = a_{0,u} + a_{1,u} \times \psi_{u,t} + a_{2,u} \times \psi_{u,t}^{2} + a_{3,u} \times \psi_{u,t}^{3} +  a_{4,u} \times \psi_{u,t}^{4}, 
\end{eqnarray}

\begin{eqnarray}\label{meq15}
	fcj_{u,t} = b_{0,u} + b_{1,u} \times ( \sum_{j=1}^{J_u} Qt_{u,t} + Qv_{u,t}) + b_{2,u} \times (\sum_{j=1}^{J_u} Qt_{u,t} + Qv_{u,t})^2 + 
\end{eqnarray}
\[b_{3,u} \times (\sum_{j=1}^{J_u} Qt_{u,t} + Qv_{u,t})^3 +  b_{4,u} \times (\sum_{j=1}^{J_u} Qt_{u,t} + Qv_{u,t})^4, \]

\noindent in which g is the acceleration of gravity, $9.8~m\cdot s^{-2}$. To convert horsepower into megawatts we used the constant $\text{k}=(10^{-3} \times m^{-1})$. The terms $\rho0_{uj}, \cdots, \rho5_{uj}$ are operative coefficients of turbine-generator $(j)$ at HPP $(u)$. $hl_{uj,l}$ is the is net water head of unit $(j)$ at time $(t)$ in HPP $u$. $\Delta_{H_{ujt}}$ is the sum of pen-stock losses.  $Hb_{u,t}$ is hydraulic head of the reservoir and $Qt_{uj,t}$ is the water discharge of unit $(j)$ at time $(t)$. $fcm_{u,t}$ is the height upstream of HPP $u$ at time $t$. Terms $a_{0,u}, \cdots, a_{4,u}$ are the coefficients for the fourth order polynomial of HPP $(u)$ to obtain $fcm_{u,t}$.  $fcj_{u,t}$ is the height of the HPP downstream at time $(t)$, and $b_{0,u}, \cdots, b_{4,u}$ are the coefficients for the fourth order polynomial of HPP $(u)$ that defines $fcj_{u,t}$.

\begin{table*}[!ht]
	\begin{center} \scriptsize
		\caption{Modelling notation of cascade HPP }
		\label{not}
		\begin{tabular}{cl}
			\hline
			\textbf{ Parameter} &  \textbf{Description} \\
			\hline
			$u$ & is the amount of HPP in the system \\
			$U$ & is acceleration of gravity \\
			$j$ & is the HPP turbine-generetor \\
			$J_u$ & is the total turbine-generators in HPP \\
			$t$ & is the time \\
			$ph_{uj,t}$ & is the power ($MW$) generated in turbine-generator $(j)$ of HPP $(u)$ in time $(t)$ \\
			$\psi_{u,t}$ & is the reservoir volume ($hm^3$) of HPP $(u)$ in time $(t)$ \\
			$\text{c}$ & is a constant to convert the water discharge ($m^3 \times s^{-1}$) in water volume ($hm^3$) in time $(t)$ \\
			$Qa_{u,t-1}$ & is the affluent flow ($m^3 \times s^{-1}$)  that comes to reservoir of HPP $(u)$ in time $(t-1)$ \\
			$w$ & is the HPP index to means that the defluent flow comes in reservoir of HPP $(u)$ \\
			$td$  & is the time of the water needs to move of HPP $(w)$ to $(u)$ \\
			$Qt_{w,td}$  & is the turbine flow ($m^3 \times s^{-1}$)  that comes to reservoir of HPP $(u)$ in time $(td)$ from HPP $(w)$ \\
			$Qv_{w,td}$ & is the flow rate ($m^3 \times s^{-1}$) drained through the spillway from HPP $(w)$ to $(u)$ in time $(td)$ \\
			$Qt_{uj,t-1}$ & is the turbine flow ($m^3 \times s^{-1}$) used in turbine-generator $(j)$ of HPP $(u)$ in time $(t-1)$ \\
			$Qv_{u,t-1}$ & is the flow rate ($m^3 \times s^{-1}$) discharged in HPP $(u)$ in time $(t-1)$ \\
			$E_{u,t-1}$ & is the liquid evaporation (mm) over time in a day; \\ 
			$A_{u,t-1}$ & is the water area ($km^2$) occupied in reservoir of HPP $(u)$ in time $(t-1)$ \\
			$Dm_{u,t}$  & is the power demand required measured in $MW$ for HPP $(u)$ in time $(t)$ \\
			$\varepsilon$ & is the error variation (+/- 0.5\%) tolerated in the power produced in HPPs \\
			$\psi_{u}^{min}$;$\psi_{u}^{max}$  & are the volume boundaries of reservoir in HPP $(u)$\\
			$Qd_{u}^{min}$ and $Qd_{u}^{max}$ & are defluent flow boundaries of HPP $(u)$ \\
			$Qt_{uj}^{min}$ and $Qt_{uj}^{max}$ & are turbine flow boundaries in turbine-generator $(j)$ of HPP $(u)$ \\
			$Qv_{u}^{max}$ & is the maximum value for water flow rate of HPP $(u)$ \\
			$Z_{uj,t}$ & indicates the operating status of the generating unit $(j)$ at HPP $(u)$, 0 for disabled and 1 for active \\
			$\text{g}$ & is the acceleration of gravity in $9.8 m \times s^{-2}$ \\
			$\text{k}$ & is the constant to convert horsepower into megawatts $k$~($10^{-3} \times m^{-1}$);  \\
			$\rho0_{uj}, \ldots, \rho5_{uj}$ & are operatives coefficients of turbine-generator $(j)$ at HPP $(u)$ \\
			$hl_{uj,l}$ & is the is net water head of unit $(j)$ at time $(t)$ in HPP $(u)$ \\
			$\Delta_{H_{ujt}}$  & is hydraulic head of the reservoir \\
			$Hb_{u,t}$ & is the HPP turbine-generetor \\
			$fcm_{u,t}$  & is the height upstream of HPP $(u)$ at time $(t)$ \\
			$a_{0,u}, \ldots, a_{4,u}$ & are the coefficients for the fourth order polynomial of HPP $(u)$ to obtain $fcm_{u,t}$ \\
			$fcj_{u,t}$ & iis the height of the HPP downstream at time $(t)$ \\
			$b_{0,u}, \ldots, b_{4,u}$  & iare the coefficients for the fourth order polynomial of HPP $(u)$ that define $fcj_{u,t}$ \\
			\hline
		\end{tabular}
	\end{center}
\end{table*}

\subsection{Optimization modelling}

The economic dispatch of HPPs in a cascade mode is a typical optimization problem in hydro-power energy systems. In this context, most of the mathematical dispatch models in hydroelectric plants are static models, since the water balance is disregarded as the hydraulic head of the reservoir, which is an input parameter. Moreover,the volume of the reservoir is  not considered in modelling. When the water balance is incorporated in the model, naturally the problem starts to be considered as a model of a dynamic system, since the level of the reservoir changes over time. For a scenario of cascading HPPs, the water balance is essential, as there is an interference in the reservoir level of one power plant due to the influence of the flow rates of another HPP. In our proposed modelling here for cascade mode operation, the volume of the reservoir at time $t$ can be described by Equation (\ref{meq1}),
\begin{eqnarray}\label{meq1}
	\psi_{u,t} = \psi_{u,t-1} + Qa_{u,t} + Qt_{w,td} + Qv_{w,td} - Qt_{u,t-1} + 
\end{eqnarray}
\[Qv_{u,t-1} - (E_{u,t-1} \times A_{u,t-1}),\]
\noindent in which $\psi$ is the volume of a reservoir; $(u)$ and $(w)$ are HPP indexes, $(u \neq w)$; $(td)$ is the time needed to cause water displacement between $(u)$ and $(w)$; $\upsilon$ is the reservoir volume; $Qa$ is the affluent flow; $Qt$ is the turbine flow; $Qv$ is the  flow rate, $E$ is liquid evaporation and; $A$ is the area occupied by water in the reservoir. Once the evolution of the reservoir level is considered in the model, the value of the gross drop is no longer an input parameter and becomes a variable depending on the downstream and upstream quotas. Thus, the cascade dispatch model proposed in this paper is formulated as follows: \\

\noindent \textit{Objective Functions} \\

Maximize the power production \textit{(F1)}
\begin{eqnarray}\label{meq2}
	\max F_{1}(Qt_{uj,t} . . .Qt_{UJu,t}) = \frac{1}{U}\sum_{u=1}^{U}\left (\frac{ \sum_{j=1}^{J_u} ph_{uj,t}}{\sum_{j=1}^{J_u} Qt_{uj,t}} \right),
\end{eqnarray}

\noindent in which the rate of the sum of $ph_{uj,t}$ and the sum of $Qt_{uj,t}$, from Equation (\ref{meq2}), determines the amount of energy that plant $u$ is capable of producing  given a volume of water. Maximizing this function implies generating energy with a lower water flow. \\ 

Maximize the water levels in the system's reservoirs \textit{(F2)}
\begin{eqnarray}\label{meq3}
	\max F_2(Qt_{uj,t}, \ldots, Qt_{UJu,t}) = \frac{1}{U} \sum_{u=1}^{U}\frac{\psi_{u,t}}{\psi_{u}^{max}}.
\end{eqnarray}

\noindent Maintaining a high level in the reservoirs of the system increases the robustness of the system to future drought periods. At the same time, the higher the reservoir level, the higher the upstream quota will be, leading to greater energy efficiency in power generation. \\

\noindent \textit{Constraints} \\

The electric dispatch problem in OMRS scenario of HPPs is subjected to the following equality and inequality constraints: \\

\noindent (1) The first constraint refers to the water balance of the reservoir of a HPP in the system. Thus, the Equation (\ref{meq4}) models the coupling of the operation of the HPP reservoirs in the system, 

\begin{eqnarray}\label{meq4}
	\psi_{u,t} = \psi_{u,t-1} + \text{c} \left( Qa_{u,td} + Qt_{w,td} - \sum_{j=1}^{J_u} Qt_{uj,t-1} - Qv_{u,t-1} \right) - E_{u,t-1} \times A_{u,t-1},
\end{eqnarray}

\noindent in which the term $(u)$ is the identifier index of HPP. $U$ is the amount of HPP in the system and $(j)$ is the HPP turbine-generator. Term $J_u$ is the total turbine-generators in HPP and ($t$) is the interval time;  $ph_{uj,t}$ is the power ($MW$) generated in turbine-generator $(j)$ of HPP $(u)$ in time $(t)$. Term $\psi_{u,t}$ is the reservoir volume ($hm^3$) of HPP $(u)$ in time $(t)$. Constant c is used to convert the water discharge ($m^3 \times s^{-1}$) in water volume ($hm^3$) in time $(t)$. Term $Qa_{u,t-1}$ is the affluent flow ($m^3 \times s^{-1}$)  that comes to reservoir of HPP $(u)$ in time $t-1$ and $(w)$ is the HPP index to means that the defluent flow comes in reservoir of HPP $(u)$. Term $td$ is the time of the water needs to move of HPP $(w)$ to $(u)$.  $Qt_{w,td}$ is the turbine flow ($m^3 \times s^{-1}$) that comes to reservoir of HPP $(u)$ in time $(td)$ from HPP $(w)$. Term $Qv_{w,td}$ is the flow rate ($m^3 \times s^{-1}$) drained through the spillway from HPP $(w)$ to $(u)$ in time $(td)$. $Qt_{uj,t-1}$ is the turbine flow ($m^3 \times s^{-1}$) used in turbine-generator $(j)$ of HPP $(u)$ in time $(t-1)$. Term $Qv_{u,t-1}$ is the flow rate ($m^3 \times s^{-1}$) discharged in HPP $(u)$ in time $(t-1)$.  $E_{u,t-1}$ is the liquid evaporation (mm) over time in a day and  $A_{u,t-1}$ is the water area ($km^2$) occupied in reservoir of HPP $(u)$ in time $(t-1)$. \\

\noindent (2) The second constraint, provided in the Equation (\ref{meq5}), indicates that each plant in the system must deliver a power approximately equal to the requested demand,

\begin{eqnarray}\label{meq5}
	Dm_{u,t}(1-\varepsilon) \leq \sum_{j=1}^{J_u} ph_{uj,t} \leq Dm_{u,t}(1+\varepsilon),
\end{eqnarray}

\noindent in which the power demand required $Dm_{u,t}$ is measured in $MW$ for HPP $u$ in time $t$. Term $\varepsilon$ is the error variation (+/- 0.5\%) tolerated in the power produced by Brazilian HPPs. \\

\noindent (3) Equation (\ref{meq6}) shows that the third constraint limits the volume of the reservoir to an interval relative to the limits of the minimum and maximum operating quotas,
\vskip -5mm
\begin{eqnarray}\label{meq6}
	\psi_{u}^{min} \leq \psi_{u,t} \leq \psi_{u}^{max},
\end{eqnarray}

\noindent in which $\psi_{u}^{min}$ and $\psi_{u}^{max}$ are the volume boundaries of reservoir in HPP $(u)$.  \\

\noindent (4) Fourth constraint, seen in Equation (\ref{meq7}), indicates that a plant's outflow must respect a limited range. These limits work as controls to prevent floods in regions on the river downstream from the HPP, and also for the use of water for navigation and the ecosystem in the river and in its surroundings,

\begin{eqnarray}\label{meq7}
	Qd_{u}^{min} \leq Qv_{u,t} + \sum_{j=1}^{J_u} Qt_{uj,t} \leq Qd_{u}^{max},
\end{eqnarray}

\noindent in which $Qd_{u}^{min}$ and $Qd_{u}^{max}$ are defluent flow boundaries of HPP $(u)$. $Qd_{u}^{min}$ and $Qd_{u}^{max}$ are defluent flow boundaries of HPP $(u)$.\\

\noindent (5) Fifth constraint,  Equation (\ref{meq8}) states that the turbine flows must respect the capacity limits of their respective generating units, 

\begin{eqnarray}\label{meq8}
	Qt_{uj}^{min} \leq Qt_{uj,t} \leq Qt_{uj}^{max},
\end{eqnarray}

\noindent in which $Qt_{uj}^{min}$ and $Qt_{uj}^{max}$ are turbine flow boundaries in turbine-generator $(j)$ of HPP $(u)$. $Qv_{u}^{max}$ is the maximum value for water flow rate of HPP $(u)$. \\

\noindent (6) Sixth constraint imposes a maximum limit for the flow according to Equation (\ref{meq99}), 
\vskip -10mm
\begin{eqnarray}\label{meq99}
	Qv_{u,t} \geq Qv_{u}^{max},
\end{eqnarray}

\noindent in which $Qv_{u,t}$ is the flow rate ($m^3 \times s^{-1}$) discharged in HPP $(u)$ in time $(t)$. $Qv_{u}^{max}$ is the maximum value for water flow rate of HPP $(u)$. \\

\noindent (7) Seventh constraint, seen in Equation (\ref{meq9}), states that if the volume of the reservoir exceeds its maximum operating limit, the excess water must be eliminated by the spillway. This constraint imposes a maximum limit for the flow rate,
\vskip -5mm
\begin{eqnarray}\label{meq9}
	\psi_{u,t} > \psi_{u}^{max} \Rightarrow Qv_{u,t} \geq \frac{1}{\text{c}}(\psi_{u,t} - \psi_{u}^{max} ),
\end{eqnarray}

\noindent in which $\psi_{u,t}$ is the reservoir volume ($hm^3$) of HPP $(u)$ in time $(t)$. Term $\psi_{u}^{max}$ is the maximum volume bounder of reservoir in HPP $(u)$. $Qv_{u,t-1}$ is the flow rate ($m^3 \times s^{-1}$) discharged in HPP $(u)$ in time $(t-1)$. Term c is a constant to convert the water discharge ($m^3 \times s^{-1}$) in water volume ($hm^3$) in time $(t)$. \\

\noindent (8) Eighth constraint indicates that the power generated must also respect the limits of the capacity of its generating unit,
\vskip -5mm
\begin{eqnarray}\label{meq10}
	ph_{uj}^{min} \times Z_{uj,t} \leq ph_{uj,t} \leq ph_{uj}^{max} \times Z_{uj,t}, 
\end{eqnarray}

\noindent in which $ph_{uj,t}$ is the power ($MW$) generated in turbine-generator $(j)$ of HPP $(u)$ in time $(t)$. Terms $ph_{uj}^{min}$ and $ph_{uj}^{max}$ are the boundaries of power generation. $Z_{uj,t}$ indicates the operating status of the generating unit $j$ at HPP $u$, 0 for disabled and 1 for active. \\

\noindent (9) Ninth constraint indicates that generating units have only one operating zone,
\vskip -10mm
\begin{eqnarray}\label{meq11}
	Z_{uj,t} \in \{0,1\}. 
\end{eqnarray}

\noindent  in which $Z_{uj,t}$ indicates the operating status of the generating unit $(j)$ at HPP $(u)$, 0 for disabled and 1 for active.

To satisfy these nine constrains, we apply a penalty factor ($p$) to the objective functions $F1$ and $F2$. Thus, the fitness functions applied of the OMRS problem are defined by Equations (\ref{fit1}) and (\ref{fit2}), according to:

\begin{eqnarray}\label{fit1}
	F'_1= p \sum^{n}_{i} \max \left[0,\frac{ \sum_{j=1}^{J_u} ph_{uj,t}}{\sum_{j=1}^{J_u} Qt_{uj,t}}\right]^2, ~~and
\end{eqnarray}

\begin{eqnarray}\label{fit2}
	F'_2= p \sum^{n}_{i} \max \left[0, \frac{\psi_{u,t}}{\psi_{u}^{max}}\right]^2, ~~  where ~~p=0.5.
\end{eqnarray}

\section{Experiments and results}

This section presents the experimental results of the paper. We have structured the experiments carried out in two different parts: first, a set of experiments on continuous benchmark functions analyzes the performance of the proposed MESH algorithm in well-known problems, and we have used these results to set the best configuration  of the algorithm. Then, we have tested the MESH approach in a real problem of electric dispatch problem in a cascade operation with multiple reservoirs, comparing the results obtained with other state-of-the-art MOEAs.

\subsection{Evaluation of the MESH performance in continuous benchmark functions}\label{exper}

In this section, the experimental performance of the MESH algorithm in some well-known continuous benchmark problems is analyzed. The goal is twofold: (i) to determine the best algorithm configuration considering the problem sets, and (ii) to compare the best version of MESH with four algorithms (NSGA-II, SPEA-2, MOEA/D, and NSGA-III) for solving the problems. Thus, the experimental setup is divided into the following case studies:

\begin{enumerate}
	\item to determine the best algorithm configuration for MESH, we use a well-known set of benchmark functions. Several algorithm configurations are employed to solve the problems and a statistical inference is applied to determine the best configuration.
	\item to verify the MESH performance a preliminary experiment is conducted. We use the same set of benchmark functions to compare our algorithm with the standard algorithms (NSGA-II, SPEA-2, MOEA/D, and NSGA-III). For that, statistical inference techniques have been adopted, such as: analysis of variance (ANOVA) and, multiple comparison test (Tukey) as described in \citep{ mont}.
\end{enumerate}
In all experiments, the best non-dominated set of last generation and the hypervolume are used as an indicator for assessing the algorithm's performance. We performed the computational simulation using an AMD Ryzen 7 3700X with CPUs@3.60 GHz and 32 GB RAM, with Arch Linux operating system. The MESH code was implemented in Python 3.9 language programming.

\subsubsection{Algorithm configuration} \label{finetuning}

In this experiment, we aim to identify a good algorithm configuration for MESH such as to choose the particle guide, the sampling vector and the mutation strategy. The particle guide can be chosen according to the following: A particle from memory (E1) and a particle close to the upper bound to the actual Pareto front (E2). The three sampling vectors can be: swarm (V1); memory (V2) and a combination between V1 and V2 generating the (V3). We have tested the following mutation strategies' options (taken from Differential Evolution algorithm): DE/Rand/1/bin (D1); DE/Rand/2/bin (D2); DE/Best/1/Bin (D3); DE/Current-to-best/1/bin (D4); DE/Current-to-rand/1/bin (D5). In this way, 30 ($2 \times 3 \times 5$) different MESH configurations have been analysed.

\begin{figure*}[!ht]
	\begin{center}
		\includegraphics[scale=0.65]{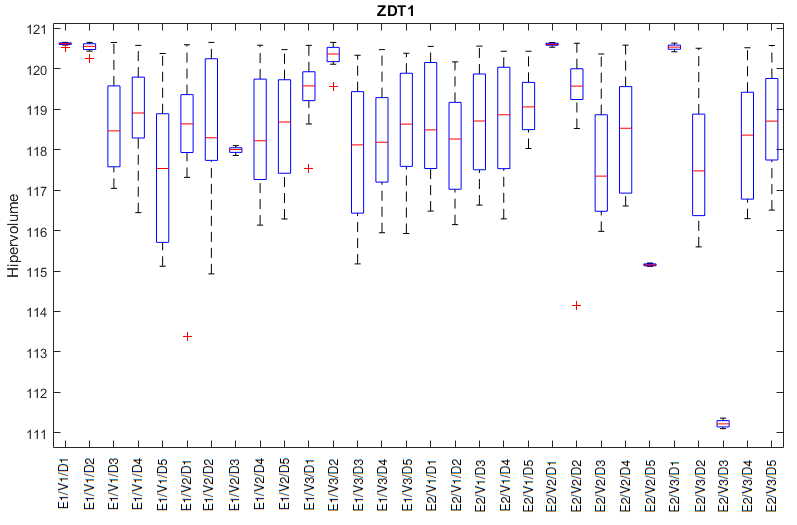}
		\caption{Boxplot of MESH versions in the ZDT1 function.} \label{boxplotzdt1}
	\end{center}
\end{figure*}

As an example, one possible setting of MESH could be E2/V1/D1, meaning the swarm guide is chosen by the Pareto Front solution, the vector selected for mutation will be from the memory and differential mutation is done with sampled vectors of the swarm population and memory under the DE/rand/1/bin strategy. Each algorithm's configuration is run 30 times using the well-known Ziztler's benchmark functions (ZTD1, ZDT2, ZDT3, ZDT4 and ZDT6) \citep{spea2} and, using the hypervolume as the performance indicator. The statistical protocol as described in \citep{carolapp} is applied. The ANOVA and the Tukey-test \citep{mont} are performed. As an example, the boxplot for all algorithms' run using ZDT1 can bee seen in Figure \ref{boxplotzdt1}.

\begin{figure*}[!ht]
	\begin{center}
		\includegraphics[scale=0.66]{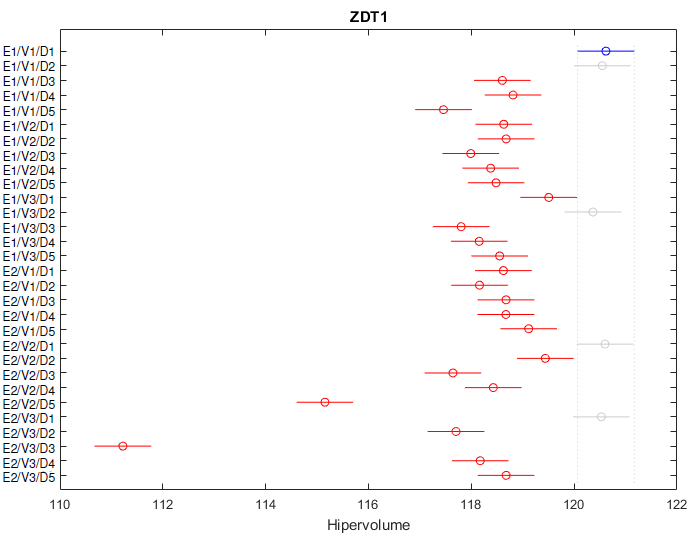}
		\caption{Tukey test of MESH versions in the ZDT1 function.} \label{tukeyzdt1}
	\end{center}
\end{figure*}

Visually, since there are boxes that do not overlap, statistical differences can be identified. An ANOVA test has been performed, in which the p-value obtained is lower than the significance level adopted (<0.05), indicating that there is a difference among the means of the hyper-volume. Tukey's test has been conducted to identify the differences among the samples. Figure \ref{tukeyzdt1} shows the result obtained.
From the results we see that five versions of MESH stand out from the others. Since the higher the hypervolume values, the best is the algorithm performance, E1/V1/D1 and E2/V2/D1 configurations are the best ones. A similar behavior has been observed in the other ZDT functions. So, for the remaining tests, only  E1/V1/D1 and E2/V2/D1 versions of MESH have been  applied.

\subsubsection{Performance assessment}\label{preex}

To validate the proposed MESH algorithm, we have performed a set of tests using the ZDT and DTLZ benchmark functions. The functions ZDT1, ZDT2, ZDT3, ZDT4, ZDT6, DTLZ1, DTLZ2, DTLZ4, and DTLZ7 \citep{spea2, dtlzs}  are employed here. The two MESH configurations, E1/V1/D1 and E2/V2/D1, are compared to the standard  NSGA-II, SPEA-2,  NSGA-III, and MOEA/D.It is worthwhile to notice that the parameters of all six algorithms, such as mutation and crossover rates, have not been fine-tuned. Since the main goal of this experiment is to validate the proposed approaches, no fine-tunning of the parameters has been done. In absence of a more informed choices,  we have used the usual values found in the literature. 

\begin{figure*}[!ht]
	\begin{center}
		\includegraphics[scale=1.3]{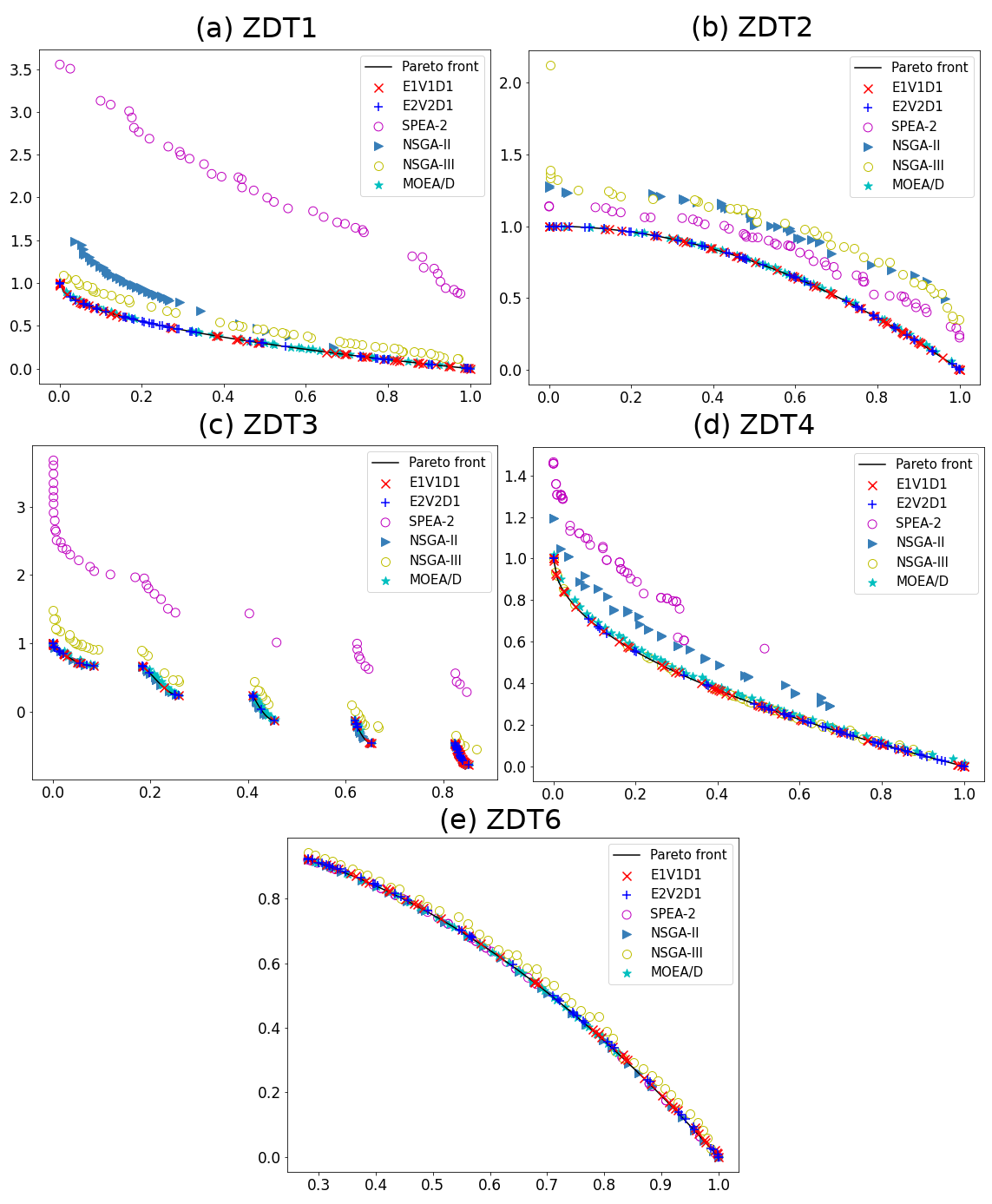}
		\caption{ Combined Pareto front for MESH versions, SPEA-2, NSGA-II, NSGA-III, and MOEA/D  for the problem tests described in  \cite{ethz}. In all graphs, the horizontal axis represents the objective function ``F1'' while the vertical axis is the objective function ``F2''. The analytical Pareto front is labelled as ``Baseline''. The number of decision variables has been set to 5 for all functions. \label{f3}}  
	\end{center}
\end{figure*}

\begin{figure*}[!ht]
	\begin{center}
		\includegraphics[scale=1.4]{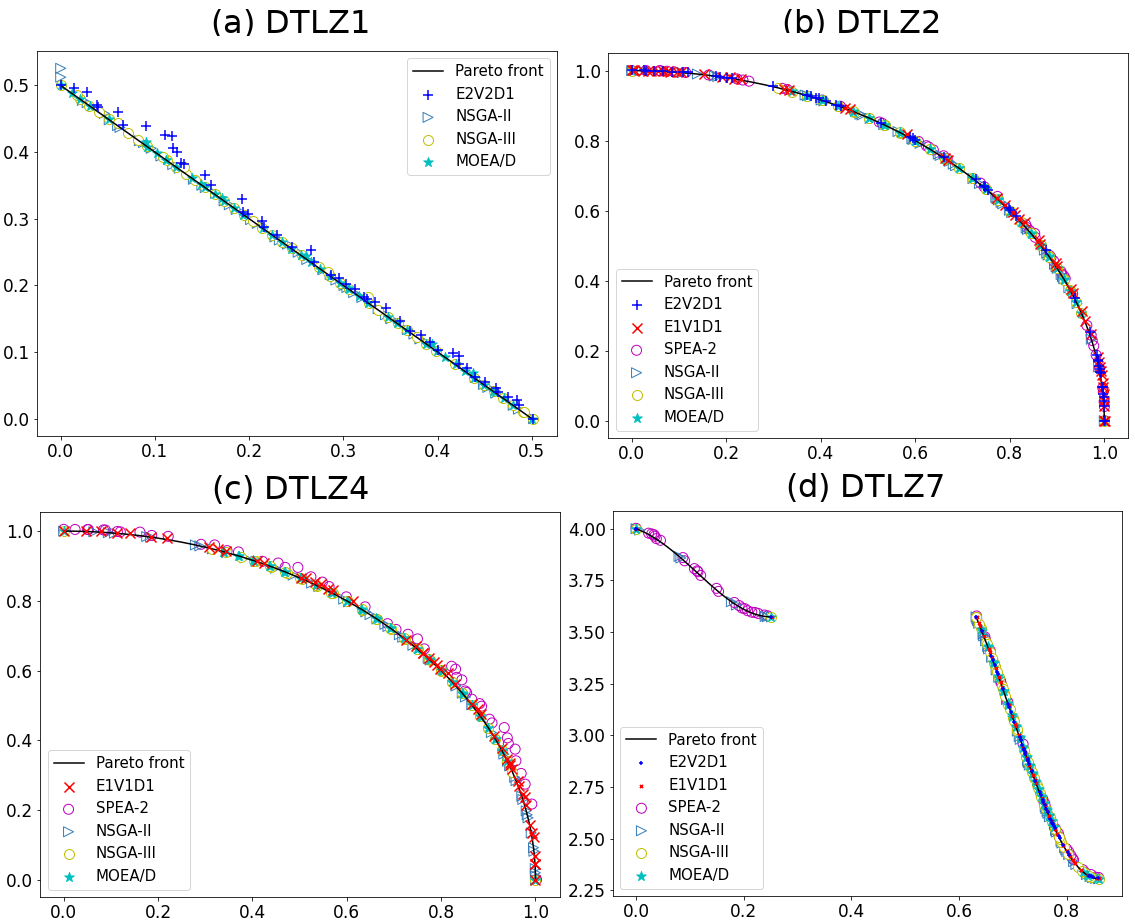}
		\caption{Combined Pareto front for MESH versions, SPEA-2, NSGA-II, NSGA-III, and MOEA/D  for the problem tests described in  \cite{dtlzs}. In all graphs, the horizontal axis represents the objective function ``F1'' while the vertical axis is the objective function ``F2''.  The analytical Pareto front is labelled as ``Baseline''. The number of decision variables has been set to 10 for all functions. \label{f4}}  
	\end{center}
\end{figure*}

For all problems, the algorithm population is set to 50 solutions. This value is also valid for the main and secondary populations such as memory, copy, file, or offspring. The parameter's values used in the algorithms are indicated in Table \ref{pinit}. Each algorithm is run 30 times. Figures \ref{f3} and \ref{f4} show the combined Pareto fronts for both MESH versions (E1/V1/D1 and E2/V2/D1), NSGA-II, SPEA-2, NSGA-III, and MOEA/D, for the ZDTs and DTLZs functions, respectively. The analytical Pareto front  of each problem is also showed. 

\begin{table*}[!ht]
	\begin{center}\small{   
			\caption{ Parameter initialization used in the algorithms considered.} \label{pinit}
			\begin{tabular}{lccccc}
				\hline
				&       MESH &    NSGA-II &      SPEA2 &   NSGA-III &     MOEA/D \\
				\hline
				Mutation rate &        0.9 &       0.02 &       0.02 &       0.02 &       0.02 \\
				
				Crossover rate &        0.7 &        0.8 &        0.7 &        0.8 &        0.7 \\
				
				Guide size &          3 &          - &          - &          - &          - \\
				
				Memory size &          5 &          - &          - &          - &          - \\
				
				Direction function &          - &          - &          - & Das-Dennis & Das-Dennis \\
				
				\hline
				Fitness evaluation &                                     \multicolumn{ 5}{c}{15000} \\
				\hline
			\end{tabular}  
			
		} 
	\end{center}
\end{table*}


Observing Figure \ref{f3}, note that both MESH versions show to be competitive when compared with the other algorithms in all functions. MESH (in both versions) finds Pareto solutions very close to the true Pareto front in functions ZDT1 and ZDT2, with better visual results than the other three algorithms (SPEA-2, NSGA-II and NSGA-III). Moreover, in ZDT3 and ZDT4, both versions of MESH visually obtain  better results when compared to  SPEA2 and NSGA-III (in ZDT3), and SPEA2 and NSGA-II (in ZDT4). In ZDT6, a visual analysis is not trivial to perform.  On the other hand, in Figure \ref{f4}, we can see that the tested algorithms are close to the analytical Pareto front. In DTLZ1 only the E2V2D1 version finds a competitive set of solutions. In the DTLZ4 function, only  E1V1D1 provides solutions and, visually, SPEA2 presents the set furthest away from the analytical result.

Graphical analysis can be a good indicator of the results obtained by the algorithms. However, a statistical test needs to be done to compare the algorithms' performance. Using the hypervolume as a performance index, an ANOVA is applied to compare the algorithms throughout the 30 runs of the algorithms. If ANOVA states there is a statistical difference between the hypervolume means of the algorithms, the Tuckey test is applied to simultaneously assess all pairwise comparisons and to identify any difference between two means that are greater than the expected standard error \cite{mont}. Table \ref{tfun} shows the hypervolume results (mean and standard deviation). The algorithm with the superior performance is indicated in bold for each problem.

\begin{table*}[!ht]
	\begin{center}\scriptsize{   
			\caption{ Hypervolume analysis using ZDT and DTLZ  functions (labelled as F). The analytical values (AN) are extracted from \cite{ethz} with reference point equal to (11,11). Legend: mean (m); standard deviation (std). } \label{tfun}
			
			\begin{tabular}{lllllllll}

				\hline
				&          F &         AN &     E1V1D1 &     E2V2D1 &    NSGA-II &   NSGA-III &      SPEA2 &     MOEA/D \\
				\hline
				&            &            &   m /(std) &   m /(std) &   m /(std) &   m /(std) &   m /(std) &   m /(std) \\
				
				& \multicolumn{ 1}{l}{1} & \multicolumn{ 1}{c}{120.657} &    \textbf{120.652} &    120.652 &     98.033 &    117.177 &     53.805 &    120.541 \\
				
				& \multicolumn{ 1}{l}{} & \multicolumn{ 1}{c}{} &    \textbf{(0.001)} &    (0.002) &      (1.929) &      (1.053) &      (2.012) &    (0.037) \\
				
				& \multicolumn{ 1}{c}{2} & \multicolumn{ 1}{c}{120.324} &    119.632 &    118.599 &     80.217 &    114.810 &     15.281 &    \textbf{120.124} \\
				
				& \multicolumn{ 1}{c}{} & \multicolumn{ 1}{c}{} &      (2.565) &      (3.846) &      (1.979) &      (1.203) &      (1.143) &      \textbf{(0.077)} \\
				
				ZDT's & \multicolumn{ 1}{c}{3} & \multicolumn{ 1}{c}{128.773} &    128.549 &    \textbf{128.661} &    101.855 &    119.808 &     55.112 &    128.372 \\
				
				& \multicolumn{ 1}{c}{} & \multicolumn{ 1}{c}{} &      (0.311) &      \textbf{(0.110)} &      (2.701) &      (0.768) &      (2.679) &      (0.169) \\
				
				& \multicolumn{ 1}{c}{4} & \multicolumn{ 1}{c}{120.657} &    \textbf{120.652} &    119.524 &    111.980 &    116.448 &     72.090 &    114.177 \\
				
				& \multicolumn{ 1}{c}{} & \multicolumn{ 1}{c}{} &      \textbf{(0.003)} &      (2.742) &      (1.919) &      (1.943) &      (3.625) &      (3.956) \\
				
				& \multicolumn{ 1}{c}{6} & \multicolumn{ 1}{c}{117.511} &    117.503 &    117.504 &    111.341 &    116.538 &    \textbf{120.287} &    102.447 \\
				
				& \multicolumn{ 1}{c}{} & \multicolumn{ 1}{c}{} &      (0.002) &      (0.001) &      (0.768) &      (0.305) &      \textbf{(0.127)} &      (2.035) \\
				\hline
				&            &            &    m (std) &    m (std) &    m (std) &    m (std) &    m (std) &    m (std) \\
				
				& \multicolumn{ 1}{l}{1} & \multicolumn{ 1}{c}{120.872} &     65.048 &    106.361 &    \textbf{120.155} &    119.941 &     60.149 &    105.345 \\
				
				& \multicolumn{ 1}{l}{} & \multicolumn{ 1}{c}{} &     (39.277) &     (2.014) &      \textbf{(1.215)} &      (1.231) &     (40.545) &     (2.501) \\
				
				& \multicolumn{ 1}{l}{2} & \multicolumn{ 1}{c}{120.207} &    120.200 &    120.200 &    120.204 &    120.204 &    120.189 &    \textbf{120.206} \\
				
				DTLZ's & \multicolumn{ 1}{l}{} & \multicolumn{ 1}{c}{} &      (0.001) &      (0.001) &      (0.001) &      (0.002) &      (0.004) &          \textbf{(0.000)} \\
				
				& \multicolumn{ 1}{l}{4} & \multicolumn{ 1}{c}{120.207} &    \textbf{119.777} &          - &    115.442 &    114.418 &    116.463 &    115.443 \\
				
				& \multicolumn{ 1}{l}{} & \multicolumn{ 1}{c}{} &     \textbf{(0.316)} &         -   &      (5.091) &      (5.082) &      (4.918) &      (5.091) \\
				
				& \multicolumn{ 1}{l}{7} & \multicolumn{ 1}{c}{116.089} &     88.816 &     88.449 &     92.481 &     92.477 &     \textbf{94.006} &     88.144 \\
				
				& \multicolumn{ 1}{l}{} & \multicolumn{ 1}{c}{} &      (6.348) &      (7.222) &      (4.846) &      (4.856) &      \textbf{(0.541)} &      (6.499) \\
				\hline
				
			\end{tabular}  
			
		} 
	\end{center}
\end{table*}  

The mean and standard deviation values are preliminary measures, but in many cases, they are not sufficient for a more effective analysis of the results. Thus, an ANOVA test is once again  performed, aiming to find possible differences between the means. Using a significance value of 5\%,  a p-value below 0.05 is found indicating that there is a difference among the means. Thus, a Tukey-test is carried out, to identify where the  differences between the samples are.

The ranking provided by the Tukey-test is also shown in Table \ref{tukeyfun}. The results indicate that both MESH versions obtain competitive results since they are classified together with the MOEA/D algorithm, in the first place, in ZDT1, ZDT2, and ZDT3. MESH is the best algorithm in ZDT4, and ties with NSGA-II, NSGA-III, and SPEA2 in ZDT6.

Regarding the DTLZ functions we can note that the E2V2D1 version of MESH is as efficient as NSGA-II, NSGA-III, and MOEA/D  in  DTLZ1. In DTLZ2, the MOEA/D has the better result when compared to others. MESH, with E1V1D1, shows a significant difference in relation to the others in DTLZ4. And finally, in DTLZ7, SPEA2 is more efficient covering a better set of solutions for this problem. Therefore, MESH is able to get significant results in six of nine known benchmark functions tested in this work. In a general analysis, it is possible to say that MESH is a competitive algorithm when applied to solve continuous problems like the ZDT and DTLZ benchmark functions. 

\begin{table*}[!ht]
	\begin{center}\scriptsize{
			\caption{ Tukey-test results using Hypervolume analysis of ZDT Benchmark Functions (labelled as F). Legend: mean (m); standard deviation (std).   \label{tukeyfun}}
			\begin{tabular}{llllll}
				\hline
				&            &    \multicolumn{ 4}{c}{Tukey-test/classification} \\
				
				&          F & 1\textsuperscript{\underline{o}} & 2\textsuperscript{\underline{o}} & 3\textsuperscript{\underline{o}} & 4\textsuperscript{\underline{o}} \\
				\hline
				& \multicolumn{ 1}{c}{1} & E1/V1/D1, E2/V2/D1 &   NSGA-III &    NSGA-II &      SPEA2 \\
				
				& \multicolumn{ 1}{c}{} &     MOEA/D &            &            &            \\
				
				& \multicolumn{ 1}{c}{2} & E1/V1/D1, E2/V2/D1 &   NSGA-III &    NSGA-II &      SPEA2 \\
				
				& \multicolumn{ 1}{c}{} &     MOEA/D &            &            &            \\
				
				& \multicolumn{ 1}{c}{3} & E1/V1/D1, E2/V2/D1 &   NSGA-III &    NSGA-II &      SPEA2 \\
				
				ZDT's & \multicolumn{ 1}{c}{} &     MOEA/D &            &            &            \\
				
				&          4 & E1/V1/D1, E2/V2/D1 &     MOEA/D &   NSGA-III & NSGA-II, SPEA2 \\
				
				&          6 &     MOEA/D & E1/V1/D1, E2/V2/D1 &   NSGA-III & NSGA-II, SPEA2 \\
				\hline
				& \multicolumn{ 1}{c}{1} & E2/V2/D1, MOEA/D & E1/V1/D1, SPEA2 &            &            \\
				
				& \multicolumn{ 1}{c}{} & NSGA-III, NSGA-II &            &            &            \\
				
				& \multicolumn{ 1}{c}{2} &     MOEA/D & NSGA-II, NSGA-III &  E1/V1/D1, &            \\
				
				& \multicolumn{ 1}{c}{} &            &            &   E2/V2/D1 &      SPEA2 \\
				
				DTLZ's & \multicolumn{ 1}{c}{4} & \multicolumn{ 1}{l}{E1/V1/D1} & MOEA/D, NSGA-III &            &            \\
				
				& \multicolumn{ 1}{c}{} & \multicolumn{ 1}{l}{} & NSGA-II, SPEA2 &            &            \\
				
				& \multicolumn{ 1}{c}{7} & \multicolumn{ 1}{l}{SPEA2} & E1/V1/D1, E2/V2/D1 &            &            \\
				
				& \multicolumn{ 1}{c}{} & \multicolumn{ 1}{l}{} & MOEA/D, NSGA-II &            &            \\
				
				& \multicolumn{ 1}{c}{} & \multicolumn{ 1}{l}{} &   NSGA-III &            &            \\
				\hline
			\end{tabular}

		}
	\end{center}
\end{table*}

\subsection{Electric dispatch simulation in cascade HPPs -- an OMRS scenario}\label{realp}

In this section we analyze the performance of the proposed MESH algorithm in a real problem of electric dispatch problem in a cascade operation, with multiple reservoirs. The same experimental methodology described in Section \ref{exper} is employed in this case. The MESH configurations, E1/V1/D1 and E2/V2/D1, are compared to standard NSGA-II, SPEA2, MOEA/D, and NSGA-III versions.  Thus, the experimental setup is divided into the following parts: 

\begin{enumerate}
	\item to assess the MESH performance to solve the electric dispatch problem in a cascade operation with multiple reservoirs, we performed the simulation model. The proposed meta-heuristic is compared with the other algorithms. The algorithms have been constructed taking into account the structures and characteristics of the real application problem studied in this work. The obtained results are analyzed using a statistical inference comparison of the results obtained by MESH versus the others using the same methodology proposed in preliminary experiment, and
	\item to analyze the results found by MESH solving the electrical dispatch problem, highlighting the positive impact of using MESH as a power production control system.
	
\end{enumerate}

\subsubsection{Simulation modelling in OMRS scenario}\label{simul}

For guiding an optimal operation of cascade reservoirs and giving full play to capacity benefits of HPP stations, the mathematical model is established based on the principles of (1) maximizing the power production and (2) maximizing the reservoir volume of cascaded HPPs. In our approach, the spatial coupling of an HPP energy system with two cascade reservoirs is made. The cascade system used for simulation is composed of an HPP ``U1'' with a maximum capacity of 528 MW/h consisting 8 turbine-generator units and another HPP ``U2'' that is downstream from U1 with a maximum capacity of 396 MW/h composed of 6 turbine-generator units installed. 

\begin{figure*}[!ht]
	\begin{center}
		\includegraphics[scale=0.5]{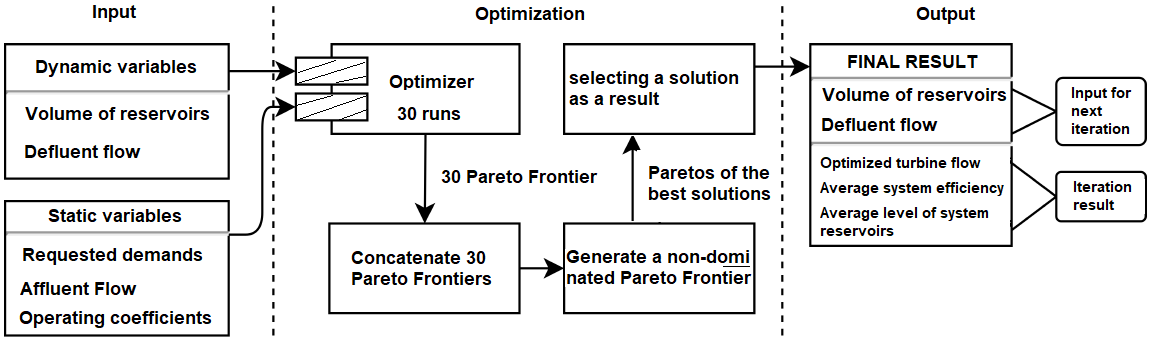}
		\caption{Flowchart for simulating an iteration of the model.\label{fsim}}
	\end{center}
\end{figure*}

The reservoirs of the two HPPs are identical and have a maximum volume of 19528 $hm^3$ and a minimum volume of 4250 $hm^3$. The initial volume for both reservoirs is 80\%, which represents a robust scenario in which there is a good availability of water in the reservoir and the height of the hydraulic head guarantees a good yield for the generating units. In this work, a restarting strategy is used to address the dynamic optimization inherent in the proposed model. Thus, whenever the model is changed over time, a new optimization is performed. The experiments carried out to validate the model have a time interval of one hour, over 24 hours as shown by Figure \ref{fsim}.

\begin{figure}[!ht]
	\begin{center}
		\includegraphics[scale=0.5]{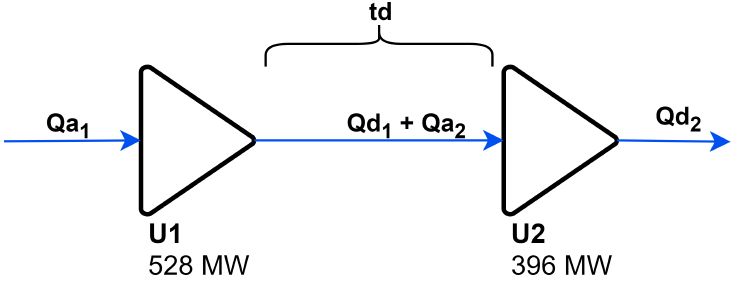}
		\caption{Diagram of the Cascade HPP system used in the simulation.\label{fsimu}}
	\end{center}
\end{figure}

In the simulation, each iteration receives two types of input variables. The set of static variables is defined before the start of the simulation and their values are independent between iterations. Dynamic variables are transmitted from one iteration to the next. From the Combined Pareto front from 30 runs, the solution more central to the set is used in the next iteration. From this solution, the states of the reservoirs and the defluent flows are transmitted to the next iteration as dynamic input variables. The dynamic power generation system adopted is shown in Figure \ref{fsimu}. As there is no other HPP downstream of U1, the terms of defluent flow ($Qt_{w,td}$ and $Qv_{w,td}$) are null for the U1 water balance. In the water balance of U2, on the other hand, the time taken to move water between U1 and U2 is $td = 2$ hours.

In this simulation system, the coefficients adopted for the upstream and downstream and efficiency production polynomials are, respectively: 
\begin{itemize}
	\item $a_0 =5.30E^{+02}$, $b_0 = 5.15E^{+02}$ and $\rho_0 = 1.46E^{-01}$; 
	\item $a_1 = 6.30E^{-03}$, $b_1 = 1.61E^{-03}$ and $\rho_1 = 1.80E^{-02}$; 
	\item $a_2=-4.84E^{-07}$, $b_2=-2.55E^{-07}$ and $\rho_2 = 5.05E^{-03}$;  
	\item $a_3=2.20E^{-11}$, $b_3=2.89E^{-11}$ and $\rho_3 = -3.52^{05}$;
	\item  $a_4=-3.84E^{-16}$, $b_4=-1.18E^{-15}$, $\rho_4 = -1.12E^{-03}$ and $\rho_5 = -1.45E^{-05}.$
\end{itemize}

The limits of the defluent flow rates are defined in the interval of [400, 2500] $m^3 \times s^{-1}$. Table \ref{dados} shows the affluent flow rate, $Qa$ ($ m^3 \times s^{-1}$) and the requested power energy demand, $Dm$ ($MW$) for both HPPs in a cascade mode operation within 24 hours.  

\begin{table}[!ht]
	\begin{center}
		\caption{Affluent flow and power demand energy data for simulation. Historical data of a critical period with low rain precipitation.}
		\label{dados}
		\begin{tabular}{ccccc|ccccc}
			\hline
			& \multicolumn{ 2}{c}{U1} & \multicolumn{ 2}{c}{U2} &            & \multicolumn{ 2}{c}{U1} & \multicolumn{ 2}{c}{U2} \\
			\hline
			$h$ &         $Qa$ &         $Dm$ &         $Qa$ &         $Dm$ &          $h$ &         $Qa$ &         $Dm$ &         $Qa$ &         $Dm$ \\
			\hline
			0 &        102 &        330 &        208 &        264 &         12 &        341 &        305 &        228 &        235 \\
			
			1 &        102 &        330 &        289 &        264 &         13 &        341 &        305 &        114 &        240 \\
			
			2 &        103 &        330 &        297 &        264 &         14 &        340 &        305 &        114 &        235 \\
			
			3 &        105 &        322 &        192 &        258 &         15 &        340 &        305 &        114 &        240 \\
			
			4 &        221 &        322 &        201 &        258 &         16 &        339 &        305 &        114 &        235 \\
			
			5 &        223 &        330 &        204 &        264 &         17 &        227 &        305 &        114 &        235 \\
			
			6 &        110 &        330 &        218 &        264 &         18 &        236 &        420 &        230 &        336 \\
			
			7 &        227 &        330 &        227 &        264 &         19 &        245 &        437 &        229 &        343 \\
			
			8 &        114 &        305 &        343 &        240 &         20 &        253 &        437 &        225 &        343 \\
			
			9 &        228 &        305 &        343 &        235 &         21 &        376 &        437 &        115 &        343 \\
			
			10 &        227 &        305 &        228 &        235 &         22 &        376 &        437 &        109 &        343 \\
			
			11 &        227 &        305 &        235 &        235 &         23 &        385 &        445 &        223 &        349 \\
			\hline
		\end{tabular}  
	\end{center}
	\vskip -10mm
\end{table}

\subsubsection{Result analysis and discussion}

\begin{landscape}
	\begin{figure*}[!ht]
		\begin{center}
			\includegraphics[scale=1]{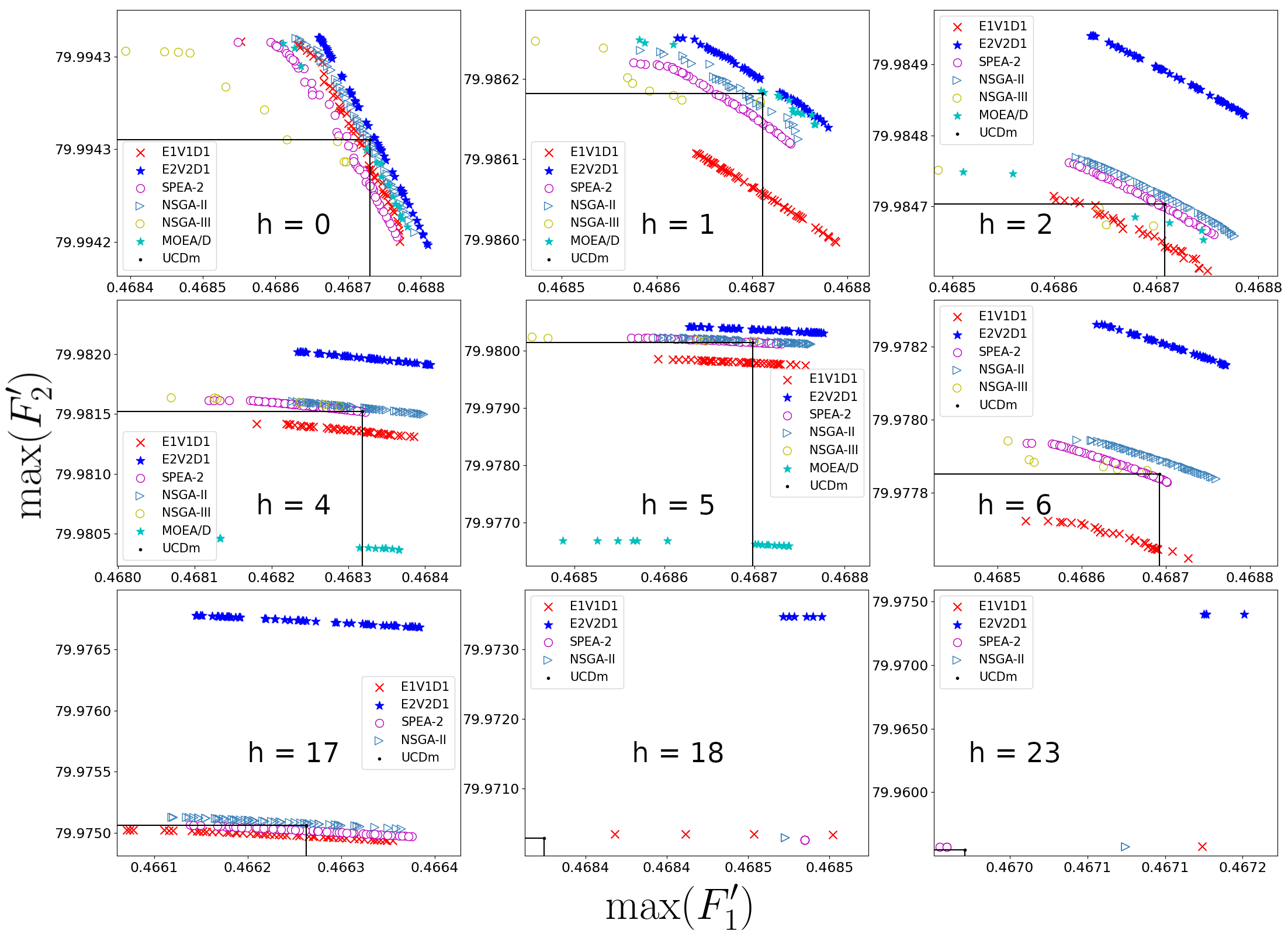}
			\caption{Combined Pareto front of the cascade model in hours: 0,1,2,4,5,6,17,18 and 23 for the MESH versions (E1V1D1 and E2V2D1), NSGA-II, SPEA-2, NSGA-III, MOEA/D. The usual usual control dispatch mode (UCDm) and the region it dominates are also indicated.  \label{modelo2}}
		\end{center}
	\end{figure*}
	
\end{landscape}

In our experimental design, the first iteration of hourly demand for each algorithm uses dynamic variables, as well as the other iterations. Each algorithm is executed 30 times. The algorithm parameters have been set as in the preliminary experiment described in subsection \ref{preex}. The Pareto Fronts are combined and the dominance operation is performed to generate a final Pareto front. The most central solution of the set is used as inputs for the next hour energy generation. Figure \ref{modelo2} shows the Combined Pareto front of some simulation hours, including the area that delimits the region dominated by the solution used for the usual control dispatch mode (UCDm, when the demand is divided equally for each turbine-generator) in HPPs.

In the first hour of the simulation, $h = 0$, the MESH with the E2V2D1 configuration is the furthest from the origin, suggesting that this configuration generates better solutions. In addition, E2V2D1 is the only algorithm that does not have any points dominated by UCDm. It is noted that NSGA-II shows a Pareto set containing a number of diverse solutions. MOEA/D presents a set of solutions that is not capable of efficiently contemplating the objective of maximizing the volume of reservoirs (F2). E1V1D1 presents a diversified Pareto set in which it visually dominates  SPEA2 and NSGA-III solutions. It is clearly noted that like MOEA/D, the NSGA-III is able to find a set of solutions that meet the conflicting goals simultaneously, the maximization of productivity (F1) and volume of reservoirs (F2).

From hour 0 to 17, the Pareto Front of the algorithms follows a pattern: solutions produced by E2V2D1 configuration are the most distant to origin, MOEA/D and NSGA-III maintain dispersed sets until 5th hour (MOEA/D) and 8th hour (NSGA-III), followed by NSGA-II, then SPEA2 and finally the E1V1D1 configuration, as exemplified  in Figure \ref{modelo2}. It is possible to notice that, after 5th hour,  MOEA/D  is no able anymore to provide solutions. After the 18th hour, due to the increase in the demand for energy in the HPPs, the feasible search space  is reduced, thus the algorithms have a greater difficulty in generating a complete Pareto set.  This fact is justified by the change in production increased by approximately 100 MW/h between 17th and 18th hours (see Table \ref{dados}). However, we emphasize that only E2V2D1, SPEA2, NSGA-II and E1V1D1 find solutions that can be used by the system dispatch control.

\begin{figure}[!ht]
	\begin{center}
		\includegraphics[scale=0.6]{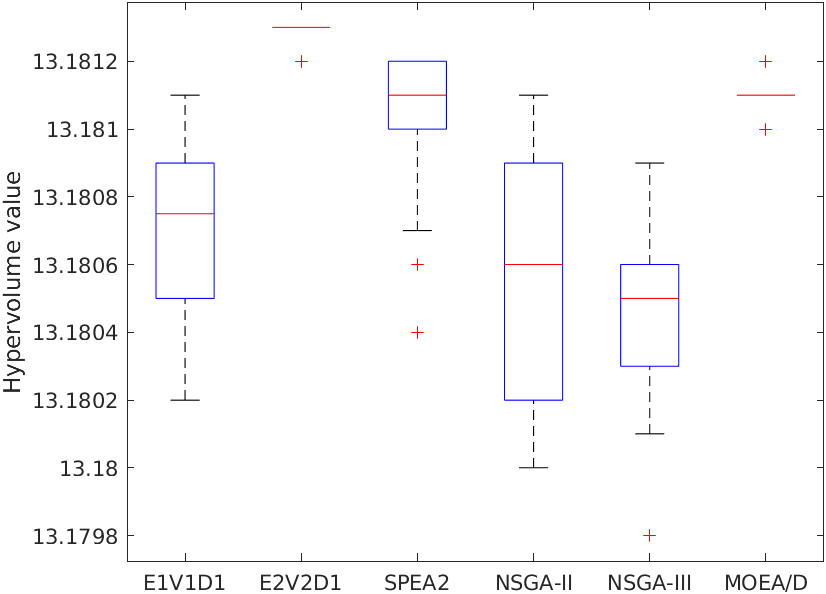}
		\caption{ Boxplot of hypervolume values of the first hour generation for all algorithms \label{bxpc}}
	\end{center}
\end{figure}

Excepting  E2V2D1, all algorithms generate solutions dominated by UCDm, in the daily control of system operation. The points found by E2V2D1 are more advantageous in terms of relation to the points of the other algorithms for keeping a high level of the reservoir. As we are proposing a new cascade dispatch model, the optimal Pareto set of this real problem is unknown. Once again, we have used the hypervolume metric \citep{spea2} to assess the algorithm performance. Note that the first hour of the simulation is the only iteration in which all the algorithms have the same initial states and, therefore, are optimizing the model under identical conditions. Figure \ref{bxpc} shows the hypervolume boxplot of the first hour generation for all algorithms. 

Boxplots are not only useful to analyze the range and distribution of the data, but sometimes it can provide information about the true difference among the means. If the notches in the boxplots do not overlap, it can be concluded, with 95\% confidence, that the true means do differ. Keeping that in mind and observing Figure \ref{bxpc}, it is possible to conclude that:

\begin{itemize}
	\item there are differences among the true means of algorithms;
	\item it is not possible to conclude if there is a statistically significant difference between the true means of E1V1D1, NSGA-II, and NSGA-III algorithms.
\end{itemize}

\begin{figure}[!ht]
	\begin{center}
		\includegraphics[scale=0.45]{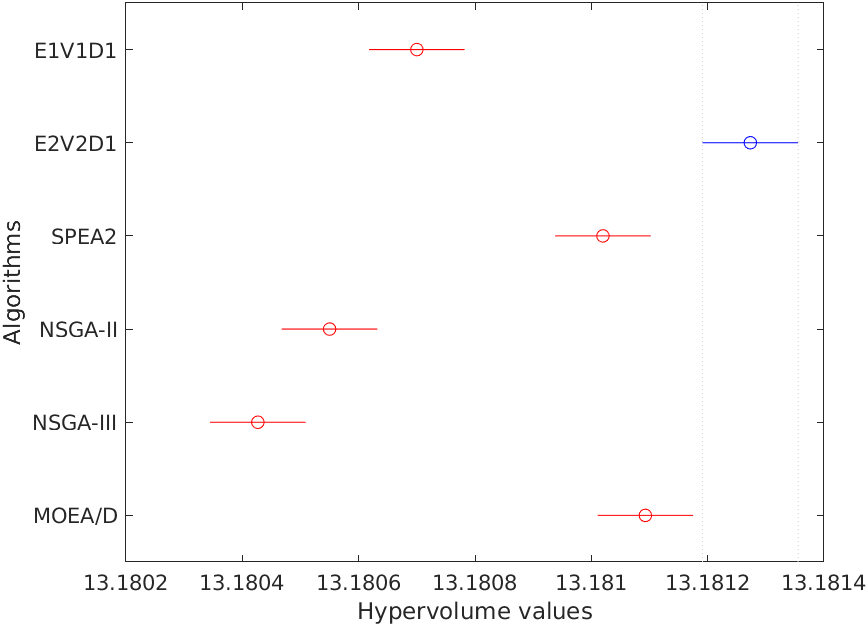}
		\caption{Cascade Tukey results\label{tikeycas}}
	\end{center}
	\vskip -5mm
\end{figure}

To statically assess the difference in performance of the tested algorithm, an ANOVA with 5\% of significance level is applied. With a p-value $<0.05$, it is possible to state there is a statistically significant difference between the algorithms' means. In sequence, Tukey test is applied indicating  which specific group's means (compared with each other) are different. Figure \ref{tikeycas}  shows the result of the Tukey test confirming that MESH in version E2V2D1 configuration generates solutions with larger hypervolume values, indicating a superior performance.

The same assessment has been made for the all hours on a daily schedule. The MESH version, E2V2D1, achieves the highest hypervolume results, differing statistically from the MOEA/D, SPEA2, E1V1D1, NSGA-II, and NSGA-III. The experimental results have showed that the proposed MESH is able to control the operation of a large multi-reservoir system producing power successfully. MESH demonstrates the effectiveness comparable or better than those presented by standard algorithms from literature. MESH has complied with all constraints imposed by the electric dispatch problem and it shows a safe approach to the operation.

\begin{figure*}[!ht]
	\begin{center}
		\includegraphics[scale=0.65]{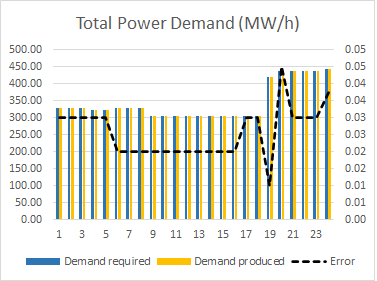} \includegraphics[scale=0.66]{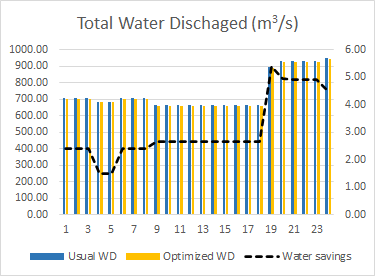}
		\caption{Electric dispatch results in HPP -- U1. The left side shows the total power generation. The right side shows the total water discharged. The line represents the small error in power production and the water savings, respectively. Legend: Usual WD (water discharge) means the operation used in usual electric dispatch control; and Optimized WD (water discharge) is the water provided by the MESH control operator. \label{thhp8}}
	\end{center}
	\vskip -5mm
\end{figure*}

Next, we aim to verify the electrical significance of MESH solutions. For that, the central Pareto Front solution for each hour is established, since this represents a compromising solution between the both objectives: (1) maximizing the power production and (2) maximizing the reservoir volume. Figure \ref{thhp8} shows the MESH ability to produce power respecting the constraints and saving water in the daily operation of HPP with eight turbine-generators. Is it important to note that this plant is operating in low demand. Even though, MESH is able to obtain optimized flows capable of saving water resources in the power production.

\begin{figure*}[!ht]
	\begin{center}
		\includegraphics[scale=0.6]{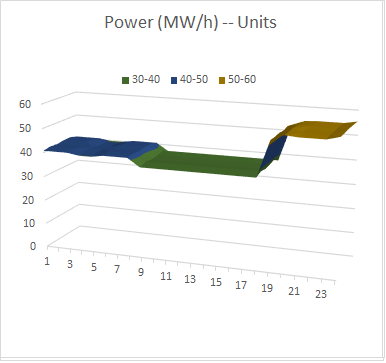} \includegraphics[scale=0.6]{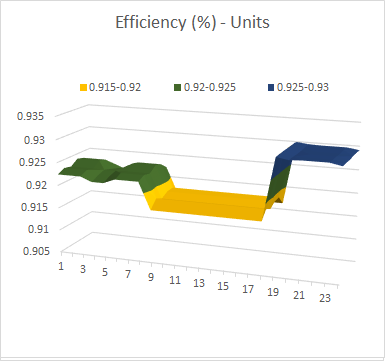}
		\caption{Power and efficiency in HPP -- U1. Left: Power production Right: Efficiency. \label{hpp8}}
	\end{center}
	\vskip -6mm
\end{figure*}

Figure \ref{hpp8} exemplifies the power generated using the eight power units (turbine-generators) and the efficiency obtained for each unit in daily electric dispatch. We can see in the Figure \ref{hpp8} that the MESH, as an electric dispatch control, is able to produce power energy respecting the boundaries since we can verify that the power generated is between 35 and 60 MW/h for each turbine-generator unit.

\begin{figure*}[!ht]
	\begin{center}
		\includegraphics[scale=0.65]{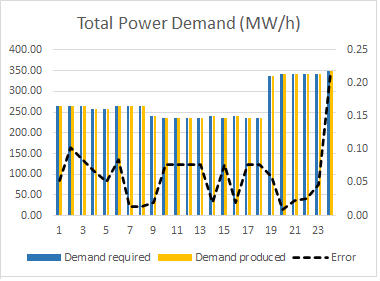} \includegraphics[scale=0.65]{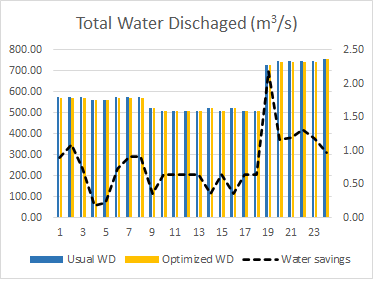}
		\caption{Electric dispatch results in HPP -- U2. The left side shows the total power generation. The right side shows the total water discharged. The line represents the small error in power production and the water savings, respectively. Legend: Usual WD (water discharge) means the operation used in usual electric dispatch control; and Optimized WD (water discharge) is the water provided by the MESH control operator.\label{thhp6}}
	\end{center}
	\vskip -5mm
\end{figure*}

It is also possible to note that the plant works in a good efficiency in which each unit reached between a 91\% and 93\% yield. We see that the closer to nominal demand, the greater the efficiency thus generating greater water savings. The central solution from Pareto Front also represents the results of the HPP that is downstream of U1. Figure \ref{thhp6} shows the total power generated and the total water discharge used by MESH in HPP -- U2. 

\begin{figure*}[!ht]
	\begin{center}
		\includegraphics[scale=0.60]{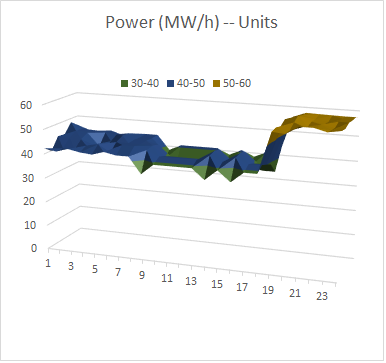} \includegraphics[scale=0.60]{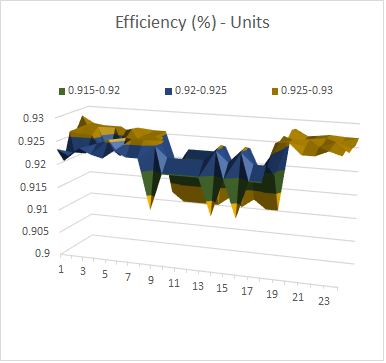}
		\caption{Power and efficiency in HPP -- U2. Left: Power production Right: Efficiency\label{hpp6}}
	\end{center}
\end{figure*}

Figure \ref{hpp6} exemplifies the power generated using the six turbine-generators and the efficiency obtained for each unit in daily electric dispatch. As we can see in the Figure \ref{hpp6}, the production behavior of the HPP that is downstream of U1 is a slightly distorted graphic.

At HPP U2, the power demanded is even lower, which for a long period the plant produced energy at 50\% of its nominal capacity. However, MESH is able to find optimized dispatches that guarantees the important constraints of the problem. We can see that all six generating units operate between 91\% and 93\% of production capacity. These optimistic results show that MESH is capable of operating plants in a Multi-Reservoir System scenario. 

MESH respects all the constraints imposed on the problem, guarantees the optimal dispatch for the cascade system, carries out the water balance maximizing the volume of water in the reservoirs, and operates the generating units at a high level of efficiency. In order to demonstrate the MESH efficiency as a control system for the electrical dispatch operation, the data report is available in Table \ref{reports}.

\begin{table*}[!ht]
	\tiny
	\begin{center} 
		\caption{Electric dispatch report in OMRS operated by MESH. Legend: Hour (h) Demand Required (DR); Demand Produced (DP); Demand error (E); Usual water discharge (UWD); Optimized Water discharge (OWD); Saved water (SW); Litters (l); Million~(mi).\label{reports}}
		\begin{tabular}{ccccccc|cccccc}
			\hline
			h &         DR &         DP &          E &        UWD &        OWD &         SW &         DR &         DP &          E &        UWD &        OWD &         SW \\
			\hline
			0 &     330.00 &     330.03 &       0.03 &     705.81 &     703.40 &       2,41 &     264.00 &     263.95 &       0.05 &     573.59 &     572.69 &       0.90 \\
			
			1 &     330.00 &     330.03 &       0.03 &     705.81 &     703.40 &       2,41 &     264.00 &     263.90 &       0.10 &     573.59 &     572.51 &       1.09 \\
			
			2 &     330.00 &     330.03 &       0.03 &     705.81 &     703.40 &       2,41 &     264.00 &     264.09 &       0.09 &     573.59 &     572.86 &       0.73 \\
			
			3 &     322.00 &     322.03 &       0.03 &     686.02 &     684.54 &       1.49 &     258.00 &     258.07 &       0.07 &     560.56 &     560.38 &       0.18 \\
			
			4 &     322.00 &     322.03 &       0.03 &     686.02 &     684.54 &       1.49 &     258.00 &     258.05 &       0.05 &     560.56 &     560.33 &       0.23 \\
			
			5 &     330.00 &     330.02 &       0.02 &     705.81 &     703.40 &       2.41 &     264.00 &     264.09 &       0.09 &     573.59 &     572.86 &       0.73 \\
			
			6 &     330.00 &     330.02 &       0.02 &     705.81 &     703.40 &       2.41 &     264.00 &     264.01 &       0.01 &     573.59 &     572.68 &       0.92 \\
			
			7 &     330.00 &     330.02 &       0.02 &     705.81 &     703.40 &       2.41 &     264.00 &     264.01 &       0.01 &     573.59 &     572.68 &       0.92 \\
			
			8 &     305.00 &     305.02 &       0.02 &     662.67 &     660.02 &       2.66 &     240.00 &     239.98 &       0.02 &     521.45 &     521.09 &       0.36 \\
			
			9 &     305.00 &     305.02 &       0.02 &     662.67 &     660.02 &       2.66 &     235.00 &     235.08 &       0.02 &     508.41 &     507.77 &       0.64 \\
			
			10 &     305.00 &     305.02 &       0.02 &     662.67 &     660.02 &       2.66 &     235.00 &     235.08 &       0.02 &     508.41 &     507.77 &       0.64 \\
			
			11 &     305.00 &     305.02 &       0.02 &     662.67 &     660.02 &       2.66 &     235.00 &     235.08 &       0.02 &     508.41 &     507.77 &       0.64 \\
			
			12 &     305.00 &     305.02 &       0.02 &     662.67 &     660.02 &       2.66 &     235.00 &     235.08 &       0.02 &     508.41 &     507.77 &       0.64 \\
			
			13 &     305.00 &     305.02 &       0.02 &     662.67 &     660.02 &       2.66 &     240.00 &     239.98 &       0.02 &     521.45 &     521.09 &       0.36 \\
			
			14 &     305.00 &     305.02 &       0.02 &     662.67 &     660.02 &       2.66 &     235.00 &     235.08 &       0.02 &     508.41 &     507.77 &       0.64 \\
			
			15 &     305.00 &     305.02 &       0.02 &     662.67 &     660.02 &       2.66 &     240.00 &     239.98 &       0.02 &     521.45 &     521.09 &       0.36 \\
			
			16 &     305.00 &     305.02 &       0.02 &     662.67 &     660.02 &       2.66 &     235.00 &     235.08 &       0.02 &     508.41 &     507.77 &       0.64 \\
			
			17 &     305.00 &     305.02 &       0.02 &     662.67 &     660.02 &       2.66 &     235.00 &     235.08 &       0.02 &     508.41 &     507.77 &       0.64 \\
			
			18 &     420.00 &     420.01 &       0.01 &     898.30 &     892.91 &       5.39 &     336.00 &     335.94 &       0.06 &     730.03 &     727.85 &       2.18 \\
			
			19 &     437.00 &     436.95 &       0.05 &     934.66 &     929.69 &       4.97 &     343.00 &     343.01 &       0.01 &     745.24 &     744.07 &       1.16 \\
			
			20 &     437.00 &     436.97 &       0.03 &     943.66 &     929.76 &       4.91 &     343.00 &     343.02 &       0.02 &     745.24 &     744.04 &       1.20 \\
			
			21 &     437.00 &     436.97 &       0.03 &     934.66 &     929.76 &       4.91 &     343.00 &     342.97 &       0.40 &     745.24 &     743.92 &       1.31 \\
			
			22 &     437.00 &     436.97 &       0.03 &     934.66 &     929.74 &       4.92 &     343.00 &     343.05 &       0.05 &     745.24 &     744.05 &       1.19 \\
			
			23 &     445.00 &     444.96 &       0.04 &     951.77 &     947.30 &       4.48 &     349.00 &     348.78 &       0.22 &     758.27 &     757.31 &       0.96 \\
			\hline
			\multicolumn{ 2}{c}{SW ($m^3/s$)} &            &            &            &            &      73.57 &            &            &            &            &            &      19.24 \\
			
			\multicolumn{ 2}{c}{SW (l/day)} &            &            &            &            &   264.8 mi &            &            &            &            &            &    69.3 mi \\
			\hline
			
		\end{tabular}  
	\end{center}
\end{table*}
From the solution obtained using MESH, the water flow savings when compared with the HPP usual control dispatch mode --UCDm -- is around 73.57 $m^3/s$ for the U1 and 19.24 $m^3/s$ in daily dispatch. Expanding these results, this is equivalent to saving approximately 264.8 million liters in U1 and 69.3 million liters in U2 of water using the optimization obtained by the MESH approach. The achieved result of energy production by MESH, in which all the turbine-generator sets work in good capacity (between 91\% and 93\%) on U1 and U2 power plants, means a percentage gain in electrical production of 0.15\% according to water savings. In practice, according to the plant's production manager, a percentage of 0.1\% generates a monthly monetary profit of \$275,000 a month. Thus, MESH can achieve a monetary profit of around \$412,500 for the cascade system providing the amount of 14.91GW at operation. The choice of the solution for this analysis is totally empirical, however such a solution exemplifies that the set of Pareto optimal solutions found by MESH is efficient in practical terms of electricity production in the Brazilian scenario.

\section{Conclusions and final remarks}\label{finalr}

In this paper we have proposed a novel hybrid algorithm for multi-objective optimization, the Multi-objective Evolutionary Swarm Hybridization -- MESH. This new optimizer can be used to address problems with conflicting or competing objectives. The guide, non-dominance and crowd distance operators are the main features introduced in MESH to make it a multi-objective algorithm, together with some novel characteristics inherit from Differential Evolution, which improves the search capabilities of the algorithm. Several tests on different benchmark problems have been conducted for choosing the best algorithm configuration for MESH.
The MESH approach, in two different versions, has shown competitive results in ZDT and DTLZ benchmark problems when compared to state-of-the-art algorithms SPEA-2, NSGA-II, MOEA/D and NSGA-III. Furthermore, results obtained after applying MESH to OMRS, a real world electrical dispatch problem, are statistically robust and indicate a superiority of MESH against other well-established MOEA's. 

Regarding the electrical dispatch in cascade mode operation, it is possible to evaluate that the proposed mathematical modelling is capable of making the generation system more efficient, with a projected water savings of around millions of liters per hour. The simulation done has showed that the MESH configurations are sensitive to the problem to be optimized. The best MESH version to solve the electric dispatch in cascade operation is the E2V2D1. Thus, when the swarm guide is obtained from a particle to the upper bound of the actual Pareto front and the sampling vector is extracted from the memory, MESH works effectively as a electric dispatch controller of cascading plants. The MESH solution is able to generate a profit of approximately \$412,500. We believe that, as a future work, a technique for choosing solutions to be used in this dynamic model can be adopted. Such an approach allows real-time decision making, so that, every hour, a Pareto solution is chosen as an input for the next generation, so that it can generate better solutions. In the end, the amount of water saved in the generation can be even larger. 

\section*{Acknowledgment}

\noindent \includegraphics[width=0.85cm]{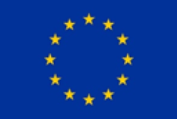} 
This project has received funding from the European Union's Horizon 2020 research and innovation programme under the Marie Sk\l{}odowska-Curie grant agreement No 754382. This research has been partially supported by Ministerio de Econom\'{i}a y Competitividad of Spain (Grant Ref. TIN2017-85887-C2-2-P) and by Comunidad de Madrid, PROMINT-CM project (grant No. P2018/EMT-4366). The authors thank UAH, UFRJ and CEFET-MG for the infrastructure used to conduct this work, and Brazilian research agencies: CAPES (Finance Code 001) and CNPq for support. ``The content of this publication does not reflect the official opinion of the European Union. Responsibility for the information and views expressed herein lies entirely with the author(s).''

\bibliography{referencias}

\begin{thebibliography}{63}
\expandafter\ifx\csname natexlab\endcsname\relax\def\natexlab#1{#1}\fi
\providecommand{\url}[1]{\texttt{#1}}
\providecommand{\href}[2]{#2}
\providecommand{\path}[1]{#1}
\providecommand{\DOIprefix}{doi:}
\providecommand{\ArXivprefix}{arXiv:}
\providecommand{\URLprefix}{URL: }
\providecommand{\Pubmedprefix}{pmid:}
\providecommand{\doi}[1]{\href{http://dx.doi.org/#1}{\path{#1}}}
\providecommand{\Pubmed}[1]{\href{pmid:#1}{\path{#1}}}
\providecommand{\bibinfo}[2]{#2}
\ifx\xfnm\relax \def\xfnm[#1]{\unskip,\space#1}\fi
\bibitem[{Almufti(2017)}]{alm}
\bibinfo{author}{Almufti, S.} (\bibinfo{year}{2017}).
\newblock \bibinfo{title}{{Using Swarm Intelligence for solving NP-Hard
  Problems}}.
\newblock {\it \bibinfo{journal}{Academic Journal of Naeroz University}\/},
  {\it \bibinfo{volume}{6}\/}, \bibinfo{pages}{46--50}.
\bibitem[{Barros et~al.(2003)Barros, Tsai, Yang \& Yeh}]{barros}
\bibinfo{author}{Barros, M.}, \bibinfo{author}{Tsai, F.},
  \bibinfo{author}{Yang, J., S.~Lopes}, \& \bibinfo{author}{Yeh, W.}
  (\bibinfo{year}{2003}).
\newblock \bibinfo{title}{Optimization of largescale hydropower system
  operations.}
\newblock {\it \bibinfo{journal}{Journal of Water Resources Planning and
  Management}\/},  {\it \bibinfo{volume}{129}\/}, \bibinfo{pages}{1878--1883}.
  \DOIprefix\doi{https://doi.org/10.1061/(ASCE)0733-9496(2003)129:3(178)}.
\bibitem[{Baumann et~al.(2017)Baumann, Peters, Weil \& Almeida}]{pes17}
\bibinfo{author}{Baumann, M.}, \bibinfo{author}{Peters, J.},
  \bibinfo{author}{Weil, C., M.~Marcelino}, \& \bibinfo{author}{Almeida, E.,
  P.~Wanner} (\bibinfo{year}{2017}).
\newblock \bibinfo{title}{Environmental impacts of different battery
  technologies in renewable hybrid micro-grids}.
\newblock {\it \bibinfo{journal}{IEEE PES ISGT-Europe}\/},  {\it
  \bibinfo{volume}{1}\/}, \bibinfo{pages}{1547--1554}.
  \DOIprefix\doi{10.1109/ISGTEurope.2017.8260137}.
\bibitem[{Cai et~al.(2001)Cai, McKinny \& Lasdon}]{cai}
\bibinfo{author}{Cai, X.}, \bibinfo{author}{McKinny, D.}, \&
  \bibinfo{author}{Lasdon, L.} (\bibinfo{year}{2001}).
\newblock \bibinfo{title}{{Solving nonlinear water management models using a
  combined genetic algorithm and linear programming approach}}.
\newblock {\it \bibinfo{journal}{Advances in Water Resources}\/},  {\it
  \bibinfo{volume}{24}\/}, \bibinfo{pages}{667--676}.
  \DOIprefix\doi{https://doi.org/10.1016/S0309-1708(00)00069-5}.
\bibitem[{Canto(2006)}]{canto}
\bibinfo{author}{Canto, S.} (\bibinfo{year}{2006}).
\newblock \bibinfo{title}{{Application of benders decomposition to power plant
  preventive maintenance scheduling}}.
\newblock {\it \bibinfo{journal}{European Journal of Operational Research}\/},
  {\it \bibinfo{volume}{184}\/}, \bibinfo{pages}{759--777}.
  \DOIprefix\doi{https://doi.org/10.1016/j.ejor.2006.11.018}.
\bibitem[{Catalao et~al.(2010)Catalao, Pousinho \& Mendes}]{catalao}
\bibinfo{author}{Catalao, J.}, \bibinfo{author}{Pousinho, H.}, \&
  \bibinfo{author}{Mendes, V.} (\bibinfo{year}{2010}).
\newblock \bibinfo{title}{{Scheduling of head-dependent cascaded reservoirs
  considering discharge ramping constraints and start/stop of units}}.
\newblock {\it \bibinfo{journal}{International Journal of Electric Power Energy
  Systems}\/},  {\it \bibinfo{volume}{32}\/}, \bibinfo{pages}{904--9010}.
  \DOIprefix\doi{https://doi.org/10.1016/j.ijepes.2010.01.022}.
\bibitem[{Chen et~al.(2016)Chen, Liu, Liu, Wei \& Mei}]{chen}
\bibinfo{author}{Chen, Y.}, \bibinfo{author}{Liu, F.}, \bibinfo{author}{Liu,
  B.}, \bibinfo{author}{Wei, W.}, \& \bibinfo{author}{Mei, S.}
  (\bibinfo{year}{2016}).
\newblock \bibinfo{title}{{ An Efficient MILP Approximation for the
  Hydro-Thermal Unit Commitment}}.
\newblock {\it \bibinfo{journal}{IEEE Transactions on Power Systems}\/},  {\it
  \bibinfo{volume}{31}\/}, \bibinfo{pages}{3318--3319}.
  \DOIprefix\doi{10.1109/TPWRS.2015.2479397}.
\bibitem[{Cinar et~al.(2010)Cinar, Kayakutlu \& Daim}]{cinar}
\bibinfo{author}{Cinar, D.}, \bibinfo{author}{Kayakutlu, G.}, \&
  \bibinfo{author}{Daim, T.} (\bibinfo{year}{2010}).
\newblock \bibinfo{title}{{Development of future energy scenarios with
  intelligent algorithms: case of hydro in Turkey}}.
\newblock {\it \bibinfo{journal}{Energy}\/},  {\it \bibinfo{volume}{35}\/},
  \bibinfo{pages}{1724--1729}.
  \DOIprefix\doi{https://doi.org/10.1016/j.energy.2009.12.025}.
\bibitem[{Cioffi \& Gallerano(2012)}]{gallerano}
\bibinfo{author}{Cioffi, F.}, \& \bibinfo{author}{Gallerano, F.}
  (\bibinfo{year}{2012}).
\newblock \bibinfo{title}{{ Multi-objective analysis of dam release flows in
  rivers downstream from hydropower reservoirs}}.
\newblock {\it \bibinfo{journal}{Applied Mathematical Modelling}\/},  {\it
  \bibinfo{volume}{36}\/}, \bibinfo{pages}{2868--2889}.
  \DOIprefix\doi{https://doi.org/10.1016/j.apm.2011.09.077}.
\bibitem[{Deb \& Jain(2014{\natexlab{a}})}]{nr9}
\bibinfo{author}{Deb, K.}, \& \bibinfo{author}{Jain, H.}
  (\bibinfo{year}{2014}{\natexlab{a}}).
\newblock \bibinfo{title}{An evolutionary many-objective optimization algorithm
  using reference-point-based nondominated sorting approach, part i: Solving
  problems with box constraints}.
\newblock {\it \bibinfo{journal}{IEEE Transactions on Evolutionary
  Computation}\/},  {\it \bibinfo{volume}{18}\/}, \bibinfo{pages}{577--601}.
  \DOIprefix\doi{10.1109/TEVC.2013.2281535}.
\bibitem[{Deb \& Jain(2014{\natexlab{b}})}]{nr10}
\bibinfo{author}{Deb, K.}, \& \bibinfo{author}{Jain, H.}
  (\bibinfo{year}{2014}{\natexlab{b}}).
\newblock \bibinfo{title}{An evolutionary many-objective optimization algorithm
  using reference-point based nondominated sorting approach, part ii: Handling
  constraints and extending to an adaptive approach}.
\newblock {\it \bibinfo{journal}{IEEE Transactions on Evolutionary
  Computation}\/},  {\it \bibinfo{volume}{18}\/}, \bibinfo{pages}{602--622}.
  \DOIprefix\doi{10.1109/TEVC.2013.2281534}.
\bibitem[{Deb et~al.(2002{\natexlab{a}})Deb, Pratap, Agarwal \&
  Meyarivan}]{nsga2}
\bibinfo{author}{Deb, K.}, \bibinfo{author}{Pratap, A.},
  \bibinfo{author}{Agarwal, S.}, \& \bibinfo{author}{Meyarivan, T.}
  (\bibinfo{year}{2002}{\natexlab{a}}).
\newblock \bibinfo{title}{{A Fast and Elitist Multiobjective Genetic Algorithm:
  NSGA II}}.
\newblock {\it \bibinfo{journal}{IEEE Transactions on Evolutionary
  Computation}\/},  {\it \bibinfo{volume}{6}\/}, \bibinfo{pages}{182--197}.
  \DOIprefix\doi{10.1109/4235.996017}.
\bibitem[{Deb et~al.(2002{\natexlab{b}})Deb, Thiele, Laumanns \&
  Zitzler}]{dtlzs}
\bibinfo{author}{Deb, K.}, \bibinfo{author}{Thiele, L.},
  \bibinfo{author}{Laumanns, M.}, \& \bibinfo{author}{Zitzler, E.}
  (\bibinfo{year}{2002}{\natexlab{b}}).
\newblock \bibinfo{title}{Scalable multioobjective optmization problems}.
\newblock {\it \bibinfo{journal}{Proc in IEEE Congress on Evolutionary
  Computation}\/},  {\it \bibinfo{volume}{1}\/}, \bibinfo{pages}{1--6}.
  \DOIprefix\doi{10.1109/CEC.2002.1007032}.
\bibitem[{ETHZ(2020)}]{ethz}
\bibinfo{author}{ETHZ} (\bibinfo{year}{2020}).
\newblock \bibinfo{title}{Density and approximations of -distribution for
  different testproblems}.
\newblock {\it \bibinfo{journal}{Available in https://sop.tik.ee.ethz.ch}\/}, .
\bibitem[{Feng et~al.(2017)Feng, Niu, Zhou \& Cheng}]{feng}
\bibinfo{author}{Feng, Z.-K.}, \bibinfo{author}{Niu, W.-J.},
  \bibinfo{author}{Zhou, J.-Z.}, \& \bibinfo{author}{Cheng, C.-T.}
  (\bibinfo{year}{2017}).
\newblock \bibinfo{title}{{ Multiobjective Operation Optimization of a Cascaded
  Hydropower System}}.
\newblock {\it \bibinfo{journal}{Journal of Water Resources Planning and
  Management}\/},  {\it \bibinfo{volume}{143}\/}, \bibinfo{pages}{1--11}.
  \DOIprefix\doi{https://doi.org/10.1061/(ASCE)WR.1943-5452.0000824}.
\bibitem[{Ge et~al.(2012)Ge, Zhang, Shu \& Ning}]{ge}
\bibinfo{author}{Ge, X.}, \bibinfo{author}{Zhang, Z.}, \bibinfo{author}{Shu,
  J.}, \& \bibinfo{author}{Ning, F.} (\bibinfo{year}{2012}).
\newblock \bibinfo{title}{{A multi-scenario model for mid-long term
  hydro-thermal optimal scheduling}}.
\newblock {\it \bibinfo{journal}{In: APPEEC}\/},  {\it \bibinfo{volume}{1}\/},
  \bibinfo{pages}{1--4}. \DOIprefix\doi{10.1109/APPEEC.2012.6306995}.
\bibitem[{Gu \& Wang(2020)}]{nr14}
\bibinfo{author}{Gu, Z.-M.}, \& \bibinfo{author}{Wang, G.-G.}
  (\bibinfo{year}{2020}).
\newblock \bibinfo{title}{Improving nsga-iii algorithms with information
  feedback models for large-scale many-objective optimization}.
\newblock {\it \bibinfo{journal}{Future Generation Computer Systems}\/},  {\it
  \bibinfo{volume}{107}\/}, \bibinfo{pages}{49--69}.
  \DOIprefix\doi{10.1016/j.future.2020.01.048}.
\bibitem[{Guan \& Zhang(1995)}]{guan}
\bibinfo{author}{Guan, X.}, \& \bibinfo{author}{Zhang, P.}
  (\bibinfo{year}{1995}).
\newblock \bibinfo{title}{{Nonlinear Approximaion Method in Lagrangian
  Relaxation-Based Algorithms for Hydrothermal Scheduling}}.
\newblock {\it \bibinfo{journal}{IEEE Transactions on Power Systems}\/},  {\it
  \bibinfo{volume}{10}\/}, \bibinfo{pages}{772--778}.
  \DOIprefix\doi{10.1109/59.387916}.
\bibitem[{Guedes et~al.(2015)Guedes, Vieira, Lisboa \& Saldanha}]{guedes}
\bibinfo{author}{Guedes, L.}, \bibinfo{author}{Vieira, D.},
  \bibinfo{author}{Lisboa, A.}, \& \bibinfo{author}{Saldanha, R.}
  (\bibinfo{year}{2015}).
\newblock \bibinfo{title}{{ A continuous compact model for cascaded hydro-power
  generation and preventive maintenance scheduling}}.
\newblock {\it \bibinfo{journal}{International Journal of Electrical Power \&
  Energy Systems}\/},  {\it \bibinfo{volume}{73}\/}, \bibinfo{pages}{702--710}.
  \DOIprefix\doi{https://doi.org/10.1016/j.ijepes.2015.05.051}.
\bibitem[{Hidalgo et~al.(2015)Hidalgo, Correia, Arnold, Estrocio, Barros,
  Fernandes \& Yeh}]{hidalgo}
\bibinfo{author}{Hidalgo, I.}, \bibinfo{author}{Correia, P.},
  \bibinfo{author}{Arnold, F.}, \bibinfo{author}{Estrocio, J.},
  \bibinfo{author}{Barros, R.}, \bibinfo{author}{Fernandes, J.}, \&
  \bibinfo{author}{Yeh, W.} (\bibinfo{year}{2015}).
\newblock \bibinfo{title}{Hybrid model for short-term scheduling of hydropower
  systems}.
\newblock {\it \bibinfo{journal}{Journal of Water Resources Planning and
  Management}\/},  {\it \bibinfo{volume}{141}\/},
  \bibinfo{pages}{04014062(1)--04014062(8)}.
  \DOIprefix\doi{https://doi.org/10.1061/(ASCE)WR.1943-5452.0000444}.
\bibitem[{Jiang et~al.(2020)Jiang, Wang \& Peng}]{nr5}
\bibinfo{author}{Jiang, E.-D.}, \bibinfo{author}{Wang, L.}, \&
  \bibinfo{author}{Peng, Z.-P.} (\bibinfo{year}{2020}).
\newblock \bibinfo{title}{Solving energy-efficient distributed job shop
  scheduling via multi-objective evolutionary algorithm with decomposition}.
\newblock {\it \bibinfo{journal}{Swarm and Evolutionary Computation}\/},  {\it
  \bibinfo{volume}{58}\/}, \bibinfo{pages}{100745}.
  \DOIprefix\doi{doi.org/10.1007/s11831-021-09562-1}.
\bibitem[{Krasnogor \& Smith(2005)}]{kra}
\bibinfo{author}{Krasnogor, N.}, \& \bibinfo{author}{Smith, J.}
  (\bibinfo{year}{2005}).
\newblock \bibinfo{title}{A tutorial for competent memetic algorithms model,
  taxonomy and design issues}.
\newblock {\it \bibinfo{journal}{IEEE Transactions on Evolutionary
  Computation}\/},  {\it \bibinfo{volume}{9}\/}, \bibinfo{pages}{474--488}.
  \DOIprefix\doi{10.1109/TEVC.2005.850260}.
\bibitem[{Leite et~al.(2002)Leite, Carneiro \& A.}]{leite}
\bibinfo{author}{Leite, P.}, \bibinfo{author}{Carneiro, A.}, \&
  \bibinfo{author}{A., C.} (\bibinfo{year}{2002}).
\newblock \bibinfo{title}{{ Energetic operation planning using genetic
  algorithms}}.
\newblock {\it \bibinfo{journal}{IEEE Transactions on Power Systems}\/},  {\it
  \bibinfo{volume}{17}\/}, \bibinfo{pages}{173--179}.
  \DOIprefix\doi{10.1109/59.982210}.
\bibitem[{Li et~al.(2020)Li, Lei \& Wang}]{nr3}
\bibinfo{author}{Li, J.}, \bibinfo{author}{Lei, A., H.~Alavi}, \&
  \bibinfo{author}{Wang, G.-G.} (\bibinfo{year}{2020}).
\newblock \bibinfo{title}{Elephant herding optimization: Variants, hybrids, and
  applications}.
\newblock {\it \bibinfo{journal}{Mathematics}\/},  {\it \bibinfo{volume}{8}\/},
  \bibinfo{pages}{1415}. \DOIprefix\doi{doi.org/10.3390/math8091415}.
\bibitem[{Li et~al.(2018)Li, Wang, Zhang \& Ishibuchi}]{nr7}
\bibinfo{author}{Li, K.}, \bibinfo{author}{Wang, R.}, \bibinfo{author}{Zhang,
  T.}, \& \bibinfo{author}{Ishibuchi, H.} (\bibinfo{year}{2018}).
\newblock \bibinfo{title}{Evolutionary many-objective optimization: A
  comparative study of the state-of-the-art}.
\newblock {\it \bibinfo{journal}{IEEE Access}\/},  {\it \bibinfo{volume}{6}\/},
  \bibinfo{pages}{26194--26214}. \DOIprefix\doi{10.1109/ACCESS.2018.2832181}.
\bibitem[{Li et~al.(2021)Li, Wang \& Gandomi}]{nr4}
\bibinfo{author}{Li, W.}, \bibinfo{author}{Wang, G.-G.}, \&
  \bibinfo{author}{Gandomi, A.} (\bibinfo{year}{2021}).
\newblock \bibinfo{title}{A survey of learning-based intelligent optimization
  algorithms}.
\newblock {\it \bibinfo{journal}{Archives of Computational Methods in
  Engineering}\/},  {\it \bibinfo{volume}{8}\/}, \bibinfo{pages}{1415}.
  \DOIprefix\doi{doi.org/10.1007/s11831-021-09562-1}.
\bibitem[{Ma et~al.(2021)Ma, Huang, Yang, Wang \& Wang}]{nr12}
\bibinfo{author}{Ma, L.}, \bibinfo{author}{Huang, M.}, \bibinfo{author}{Yang,
  S.}, \bibinfo{author}{Wang, R.}, \& \bibinfo{author}{Wang, X.}
  (\bibinfo{year}{2021}).
\newblock \bibinfo{title}{An adaptive localized decision variable analysis
  approach to large-scale multiobjective and many-objective optimization}.
\newblock {\it \bibinfo{journal}{IEEE Transactions on Cybernetics}\/},  {\it
  \bibinfo{volume}{99}\/}, \bibinfo{pages}{1--13}.
  \DOIprefix\doi{10.1109/TCYB.2020.3041212}.
\bibitem[{Mandal \& Chakraborty(2012)}]{mandal}
\bibinfo{author}{Mandal, K.}, \& \bibinfo{author}{Chakraborty, N.}
  (\bibinfo{year}{2012}).
\newblock \bibinfo{title}{{ Daily combined economic emission scheduling of
  hydrothermal systems with cascaded reservoirs using self organizing
  hierarchical particle swarm optimization technique}}.
\newblock {\it \bibinfo{journal}{Expert Systems with Applications}\/},  {\it
  \bibinfo{volume}{39}\/}, \bibinfo{pages}{3438--3445}.
  \DOIprefix\doi{https://doi.org/10.1016/j.eswa.2011.09.032}.
\bibitem[{Marano et~al.(2012)Marano, Rizzo \& Tiano}]{rizzo}
\bibinfo{author}{Marano, V.}, \bibinfo{author}{Rizzo, G.}, \&
  \bibinfo{author}{Tiano, F.} (\bibinfo{year}{2012}).
\newblock \bibinfo{title}{{Application of dynamic programming to the optimal
  management of a hybrid power plant with wind turbines, photovoltaic panels
  and compressed air energy storage}}.
\newblock {\it \bibinfo{journal}{Applied Energy}\/},  {\it
  \bibinfo{volume}{97}\/}, \bibinfo{pages}{849--859}.
  \DOIprefix\doi{https://doi.org/10.1016/j.apenergy.2011.12.086}.
\bibitem[{Marcelino et~al.(2018{\natexlab{a}})Marcelino, Almeida, Wanner,
  Baumann, Weil, Carvalho \& Miranda}]{carolapp}
\bibinfo{author}{Marcelino, C.}, \bibinfo{author}{Almeida, P.},
  \bibinfo{author}{Wanner, E.}, \bibinfo{author}{Baumann, M.},
  \bibinfo{author}{Weil, M.}, \bibinfo{author}{Carvalho, L.}, \&
  \bibinfo{author}{Miranda, V.} (\bibinfo{year}{2018}{\natexlab{a}}).
\newblock \bibinfo{title}{Solving security constrained optimal power flow
  problems: a hybrid evolutionary approach}.
\newblock {\it \bibinfo{journal}{Applied Intelligence}\/},  {\it
  \bibinfo{volume}{48}\/}, \bibinfo{pages}{3672--3690}.
  \DOIprefix\doi{https://doi.org/10.1007/s10489-018-1167-5}.
\bibitem[{Marcelino et~al.(2020)Marcelino, Baumann, Carvalho, Chibeles-Martins,
  Weil, Almeida \& Wanner}]{jors}
\bibinfo{author}{Marcelino, C.}, \bibinfo{author}{Baumann, M.},
  \bibinfo{author}{Carvalho, L.}, \bibinfo{author}{Chibeles-Martins, N.},
  \bibinfo{author}{Weil, M.}, \bibinfo{author}{Almeida, P.}, \&
  \bibinfo{author}{Wanner, E.} (\bibinfo{year}{2020}).
\newblock \bibinfo{title}{{A combined Optimization and Decision-Making approach
  for Battery-Supported HMGS}}.
\newblock {\it \bibinfo{journal}{Journal of the Operational Research
  Society}\/},  {\it \bibinfo{volume}{71}\/}, \bibinfo{pages}{762--774}.
  \DOIprefix\doi{https://doi.org/10.1080/01605682.2019.1582590}.
\bibitem[{Marcelino et~al.(2021)Marcelino, Camacho-G\'omez,
  Jim\'enez-Fern\'andez \& Salcedo-Sanz}]{energi}
\bibinfo{author}{Marcelino, C.}, \bibinfo{author}{Camacho-G\'omez, C.},
  \bibinfo{author}{Jim\'enez-Fern\'andez, S.}, \&
  \bibinfo{author}{Salcedo-Sanz, S.} (\bibinfo{year}{2021}).
\newblock \bibinfo{title}{Optimal generation scheduling in hydro-power plants
  with the coral reefs optimization algorithm}.
\newblock {\it \bibinfo{journal}{Energies}\/},  {\it
  \bibinfo{volume}{14(9)}\/}, \bibinfo{pages}{2443}.
  \DOIprefix\doi{doi.org/10.3390/en14092443}.
\bibitem[{Marcelino et~al.(2015)Marcelino, Carvalho, Almeida, Wanner \&
  Miranda}]{carolmult}
\bibinfo{author}{Marcelino, C.}, \bibinfo{author}{Carvalho, L.},
  \bibinfo{author}{Almeida, P.}, \bibinfo{author}{Wanner, E.}, \&
  \bibinfo{author}{Miranda, V.} (\bibinfo{year}{2015}).
\newblock \bibinfo{title}{Application of evolutionary multiobjective algorithms
  for solving the problem of energy dispatch in hydroelectric power plants}.
\newblock {\it \bibinfo{journal}{Lecture Notes in Computer Science}\/},  {\it
  \bibinfo{volume}{9019}\/}, \bibinfo{pages}{403--417}.
\bibitem[{Marcelino et~al.(2018{\natexlab{b}})Marcelino, Pedreira, Baumann,
  Weil, Almeida \& Wanner}]{cec18}
\bibinfo{author}{Marcelino, C.}, \bibinfo{author}{Pedreira, C.},
  \bibinfo{author}{Baumann, M.}, \bibinfo{author}{Weil, M.},
  \bibinfo{author}{Almeida, P.}, \& \bibinfo{author}{Wanner, E.}
  (\bibinfo{year}{2018}{\natexlab{b}}).
\newblock \bibinfo{title}{{Applying C-DEEPSO to Solve Large Scale Global
  Optimization Problems }}.
\newblock {\it \bibinfo{journal}{IEEE Congress on Evolutionary Computation}\/},
   {\it \bibinfo{volume}{1}\/}, \bibinfo{pages}{1547--1554}.
  \DOIprefix\doi{https://10.1109/CEC.2016.7743973}.
\bibitem[{Marcelino et~al.(2019)Marcelino, Pedreira, Baumann, Weil, Almeida \&
  Wanner}]{viab}
\bibinfo{author}{Marcelino, C.}, \bibinfo{author}{Pedreira, C.},
  \bibinfo{author}{Baumann, M.}, \bibinfo{author}{Weil, M.},
  \bibinfo{author}{Almeida, P.}, \& \bibinfo{author}{Wanner, E.}
  (\bibinfo{year}{2019}).
\newblock \bibinfo{title}{A viability study of renewables and energy storage
  systems using multicriteria decision making and an evolutionary approach}.
\newblock {\it \bibinfo{journal}{Lecture Notes in Computer Science}\/},  {\it
  \bibinfo{volume}{11411}\/}, \bibinfo{pages}{655--668}.
  \DOIprefix\doi{https://10.1007/978-3-030-12598-1\_52}.
\bibitem[{Moeini et~al.(2011)Moeini, Afshar \& Afshar}]{moe}
\bibinfo{author}{Moeini, R.}, \bibinfo{author}{Afshar, A.}, \&
  \bibinfo{author}{Afshar, M.} (\bibinfo{year}{2011}).
\newblock \bibinfo{title}{{ Fuzzy rule-based model for hydropower reservoirs
  operation}}.
\newblock {\it \bibinfo{journal}{International Journal of Electrical Power \&
  Energy Systems}\/},  {\it \bibinfo{volume}{33}\/}, \bibinfo{pages}{171--178}.
  \DOIprefix\doi{https://doi.org/10.1016/j.ijepes.2010.08.012}.
\bibitem[{Montgomery(2012)}]{mont}
\bibinfo{author}{Montgomery, D.} (\bibinfo{year}{2012}).
\newblock {\it \bibinfo{title}{{Design and analysis of Experiments}}\/}.
\newblock \bibinfo{publisher}{8th edition}.
\bibitem[{Naresh \& Sharma(2002)}]{naresh}
\bibinfo{author}{Naresh, R.}, \& \bibinfo{author}{Sharma, J.}
  (\bibinfo{year}{2002}).
\newblock \bibinfo{title}{{Short term hydro scheduling using two-phase neural
  network}}.
\newblock {\it \bibinfo{journal}{International Journal of Electrical Power \&
  Energy Systems}\/},  {\it \bibinfo{volume}{24}\/}, \bibinfo{pages}{583--590}.
  \DOIprefix\doi{https://doi.org/10.1016/S0142-0615(01)00069-2}.
\bibitem[{Niu-W-J. et~al.(2018)Niu-W-J., Feng, Cheng \& Wu}]{niu}
\bibinfo{author}{Niu-W-J.}, \bibinfo{author}{Feng, Z.-K.},
  \bibinfo{author}{Cheng, C.-T.}, \& \bibinfo{author}{Wu, X.-Y.}
  (\bibinfo{year}{2018}).
\newblock \bibinfo{title}{{ A parallel multi-objective particle swarm
  optimization for cascade hydropower reservoir operation in southwest China}}.
\newblock {\it \bibinfo{journal}{Applied Soft Computing}\/},  {\it
  \bibinfo{volume}{70}\/}, \bibinfo{pages}{562--575}.
  \DOIprefix\doi{https://doi.org/10.1016/j.asoc.2018.06.011}.
\bibitem[{Padhye et~al.(2009)Padhye, Branke \& Mostaghim}]{mostaghim}
\bibinfo{author}{Padhye, N.}, \bibinfo{author}{Branke, J.}, \&
  \bibinfo{author}{Mostaghim, S.} (\bibinfo{year}{2009}).
\newblock \bibinfo{title}{{Empirical Comparison of MOPSO Methods: Guide
  Selection and Diversity Preservation}}.
\newblock {\it \bibinfo{journal}{IEEE Congress on Evolutionary
  Computation.}\/},  {\it \bibinfo{volume}{1}\/}, \bibinfo{pages}{2516--2523}.
  \DOIprefix\doi{10.1109/CEC.2009.4983257}.
\bibitem[{Qin et~al.(2021)Qin, Sun, Jin, Tan \& Fieldsend}]{nr13}
\bibinfo{author}{Qin, S.}, \bibinfo{author}{Sun, C.}, \bibinfo{author}{Jin,
  Y.}, \bibinfo{author}{Tan, Y.}, \& \bibinfo{author}{Fieldsend, J.}
  (\bibinfo{year}{2021}).
\newblock \bibinfo{title}{Large-scale evolutionary multi-objective optimization
  assisted by directed sampling}.
\newblock {\it \bibinfo{journal}{IEEE Transactions on Evolutionary
  Computation}\/},  {\it \bibinfo{volume}{1}\/}, \bibinfo{pages}{1--15}.
  \DOIprefix\doi{10.1109/TEVC.2021.3063606}.
\bibitem[{Roefs \& Bodin(1970)}]{roefs}
\bibinfo{author}{Roefs, T.}, \& \bibinfo{author}{Bodin, L.}
  (\bibinfo{year}{1970}).
\newblock \bibinfo{title}{Multi-reservoir operation studies}.
\newblock {\it \bibinfo{journal}{Water Resources Research}\/},  {\it
  \bibinfo{volume}{6}\/}, \bibinfo{pages}{410--420}.
\bibitem[{Scuzziato et~al.(2020)Scuzziato, Finardi \& Frangioni}]{finardi}
\bibinfo{author}{Scuzziato, M.}, \bibinfo{author}{Finardi, E.}, \&
  \bibinfo{author}{Frangioni, A.} (\bibinfo{year}{2020}).
\newblock \bibinfo{title}{{Solving stochastic hydrothermal unit commitment with
  a new primal recovery technique based on Lagrangian solutions}}.
\newblock {\it \bibinfo{journal}{International Journal of Electrical Power \&
  Energy Systems}\/},  {\it \bibinfo{volume}{127}\/}, \bibinfo{pages}{1--11}.
  \DOIprefix\doi{https://doi.org/10.1016/j.ijepes.2020.106661}.
\bibitem[{Sharma et~al.(2004)Sharma, Jha \& Neresh}]{sharma}
\bibinfo{author}{Sharma, V.}, \bibinfo{author}{Jha, R.}, \&
  \bibinfo{author}{Neresh, R.} (\bibinfo{year}{2004}).
\newblock \bibinfo{title}{Optimal multi-reservoir network control by two phase
  neural network}.
\newblock {\it \bibinfo{journal}{Electrical Power and Energy Systems}\/},  {\it
  \bibinfo{volume}{68}\/}, \bibinfo{pages}{221--228}.
  \DOIprefix\doi{https://doi.org/10.1016/j.epsr.2003.06.002}.
\bibitem[{Sun et~al.(2020)Sun, Miao, Gong, Zeng, Li \& Wang}]{nr2}
\bibinfo{author}{Sun, J.}, \bibinfo{author}{Miao, Z.}, \bibinfo{author}{Gong,
  D.}, \bibinfo{author}{Zeng, X.-J.}, \bibinfo{author}{Li, J.}, \&
  \bibinfo{author}{Wang, G.-G.} (\bibinfo{year}{2020}).
\newblock \bibinfo{title}{High performance computing for cyber physical social
  systems by using evolutionary multi-objective optimization algorithm}.
\newblock {\it \bibinfo{journal}{IEEE Transactions on Emerging Topics in
  Computing}\/},  {\it \bibinfo{volume}{8}\/}, \bibinfo{pages}{20--30}.
  \DOIprefix\doi{10.1109/TETC.2017.2703784}.
\bibitem[{Tapia et~al.(2020)Tapia, Reina \& Millan}]{tapia}
\bibinfo{author}{Tapia, A.}, \bibinfo{author}{Reina, D.}, \&
  \bibinfo{author}{Millan, P.} (\bibinfo{year}{2020}).
\newblock \bibinfo{title}{{ Optimized micro-hydro power plants layout design
  using messy genetic algorithms}}.
\newblock {\it \bibinfo{journal}{Expert Systems with Applications}\/},  {\it
  \bibinfo{volume}{159}\/}, \bibinfo{pages}{1--15}.
  \DOIprefix\doi{https://doi.org/10.1016/j.eswa.2020.113539}.
\bibitem[{Wang et~al.(2020)Wang, Cai, Cui, Min \& Chen}]{nr1}
\bibinfo{author}{Wang, G.-G.}, \bibinfo{author}{Cai, X.}, \bibinfo{author}{Cui,
  Z.}, \bibinfo{author}{Min, G.}, \& \bibinfo{author}{Chen, J.}
  (\bibinfo{year}{2020}).
\newblock \bibinfo{title}{High performance computing for cyber physical social
  systems by using evolutionary multi-objective optimization algorithm}.
\newblock {\it \bibinfo{journal}{IEEE Transactions on Emerging Topics in
  Computing}\/},  {\it \bibinfo{volume}{8}\/}, \bibinfo{pages}{20--30}.
  \DOIprefix\doi{10.1109/TETC.2017.2703784}.
\bibitem[{Wang et~al.(2015)Wang, Huang, Ma \& Chen}]{jwang}
\bibinfo{author}{Wang, J.}, \bibinfo{author}{Huang, W.}, \bibinfo{author}{Ma,
  G.}, \& \bibinfo{author}{Chen, S.} (\bibinfo{year}{2015}).
\newblock \bibinfo{title}{{An improved partheno genetic algorithm for
  multi-objective economic dispatch in cascaded hydropower systems}}.
\newblock {\it \bibinfo{journal}{International Journal of Electrical Power \&
  Energy Systems}\/},  {\it \bibinfo{volume}{67}\/}, \bibinfo{pages}{591--697}.
  \DOIprefix\doi{https://doi.org/10.1016/j.ijepes.2014.12.037}.
\bibitem[{Wang et~al.(2012)Wang, Zhow, Zhou, Wang, Qin \& Lu}]{lu}
\bibinfo{author}{Wang, Y.}, \bibinfo{author}{Zhow, J.}, \bibinfo{author}{Zhou,
  C.}, \bibinfo{author}{Wang, Y.}, \bibinfo{author}{Qin, H.}, \&
  \bibinfo{author}{Lu, Y.} (\bibinfo{year}{2012}).
\newblock \bibinfo{title}{{An improved self-adaptive PSO technique for
  short-term hydrothermal scheduling}}.
\newblock {\it \bibinfo{journal}{Expert Systems with Applications}\/},  {\it
  \bibinfo{volume}{39}\/}, \bibinfo{pages}{2288--2295}.
  \DOIprefix\doi{https://doi.org/10.1016/j.eswa.2011.08.007}.
\bibitem[{Xie et~al.(2018)Xie, Wen \& Zeng}]{xie}
\bibinfo{author}{Xie, X.}, \bibinfo{author}{Wen, S.}, \& \bibinfo{author}{Zeng,
  Z.} (\bibinfo{year}{2018}).
\newblock \bibinfo{title}{{Memristor-based circuit implementation of
  pulse-coupled neural network with dynamical threshold generators}}.
\newblock {\it \bibinfo{journal}{Neurocomputing}\/},  {\it
  \bibinfo{volume}{284}\/}, \bibinfo{pages}{10--16}.
  \DOIprefix\doi{https://doi.org/10.1016/j.neucom.2018.01.024}.
\bibitem[{Xin-gang et~al.(2020)Xin-gang, Ji, Jin \& Ying}]{xinli}
\bibinfo{author}{Xin-gang, Z.}, \bibinfo{author}{Ji, L.}, \bibinfo{author}{Jin,
  M.}, \& \bibinfo{author}{Ying, Z.} (\bibinfo{year}{2020}).
\newblock \bibinfo{title}{{ An improved quantum particle swarm optimization
  algorithm for environmental economic dispatch}}.
\newblock {\it \bibinfo{journal}{Expert Systems with Applications}\/},  {\it
  \bibinfo{volume}{152}\/}, \bibinfo{pages}{1--14}.
  \DOIprefix\doi{https://doi.org/10.1016/j.eswa.2020.113370}.
\bibitem[{Yi et~al.(2018)Yi, Deb, Dong, Alavi \& Wang}]{nr6}
\bibinfo{author}{Yi, J.-H.}, \bibinfo{author}{Deb, S.}, \bibinfo{author}{Dong,
  J.}, \bibinfo{author}{Alavi, A.}, \& \bibinfo{author}{Wang, G.-G.}
  (\bibinfo{year}{2018}).
\newblock \bibinfo{title}{An improved nsga-iii algorithm with adaptive mutation
  operator for big data optimization problems}.
\newblock {\it \bibinfo{journal}{Future Generation Computer Systems}\/},  {\it
  \bibinfo{volume}{88}\/}, \bibinfo{pages}{571--585}.
  \DOIprefix\doi{doi.org/10.1016/j.future.2018.06.008}.
\bibitem[{Yi et~al.(2020)Yi, Xing, Wang, Dong, Vasilakos \& Alavi}]{nr15}
\bibinfo{author}{Yi, J.-H.}, \bibinfo{author}{Xing, l.-N.},
  \bibinfo{author}{Wang, G.-G.}, \bibinfo{author}{Dong, J.},
  \bibinfo{author}{Vasilakos, A.}, \& \bibinfo{author}{Alavi, L., A.~Wang}
  (\bibinfo{year}{2020}).
\newblock \bibinfo{title}{Behavior of crossover operators in nsga-iii for
  large-scale optimization problems}.
\newblock {\it \bibinfo{journal}{Information Sciences}\/},  {\it
  \bibinfo{volume}{509}\/}, \bibinfo{pages}{470--487}.
  \DOIprefix\doi{doi.org/10.1016/j.ins.2018.10.005}.
\bibitem[{Yoo(2009)}]{Yoo}
\bibinfo{author}{Yoo, J.-H.} (\bibinfo{year}{2009}).
\newblock \bibinfo{title}{Maximization of hydropower generation through the
  application of a linear programming model}.
\newblock {\it \bibinfo{journal}{Journal of Hydrology}\/},  {\it
  \bibinfo{volume}{376}\/}, \bibinfo{pages}{182--187}.
  \DOIprefix\doi{https://doi.org/10.1016/j.jhydrol.2009.07.026}.
\bibitem[{Zhang et~al.(2013{\natexlab{a}})Zhang, Zhou, Fang, Zhang \&
  Zhang}]{fanzou}
\bibinfo{author}{Zhang, H.}, \bibinfo{author}{Zhou, J.}, \bibinfo{author}{Fang,
  N.}, \bibinfo{author}{Zhang, R.}, \& \bibinfo{author}{Zhang, Y.}
  (\bibinfo{year}{2013}{\natexlab{a}}).
\newblock \bibinfo{title}{{ An efficient multi-objective adaptive differential
  evolution with chaotic neuron network a/nd its application on long-term
  hydropower operation with considering ecological environment problem}}.
\newblock {\it \bibinfo{journal}{International Journal of Electrical Power \&
  Energy Systems}\/},  {\it \bibinfo{volume}{45}\/}, \bibinfo{pages}{60--70}.
  \DOIprefix\doi{https://doi.org/10.1016/j.ijepes.2012.08.069}.
\bibitem[{Zhang \& Li(2007)}]{nr8}
\bibinfo{author}{Zhang, Q.}, \& \bibinfo{author}{Li, H.}
  (\bibinfo{year}{2007}).
\newblock \bibinfo{title}{Moea/d: A multiobjective evolutionary algorithm based
  on decomposition}.
\newblock {\it \bibinfo{journal}{IEEE Transactions on Evolutionary
  Computation}\/},  {\it \bibinfo{volume}{11}\/}, \bibinfo{pages}{712--731}.
  \DOIprefix\doi{10.1109/TEVC.2007.892759}.
\bibitem[{Zhang et~al.(2017)Zhang, Chen, Yao, Ba \& Ma}]{zhangg}
\bibinfo{author}{Zhang, R.}, \bibinfo{author}{Chen, D.}, \bibinfo{author}{Yao,
  W.}, \bibinfo{author}{Ba, D.}, \& \bibinfo{author}{Ma, X.}
  (\bibinfo{year}{2017}).
\newblock \bibinfo{title}{{ Non-linear fuzzy predictive control of
  hydroelectric system}}.
\newblock {\it \bibinfo{journal}{IET Generation, Transmission \&
  Distribution}\/},  {\it \bibinfo{volume}{11}\/}, \bibinfo{pages}{1966--1975}.
  \DOIprefix\doi{https://doi.org/10.1049/iet-gtd.2016.1300}.
\bibitem[{Zhang et~al.(2013{\natexlab{b}})Zhang, Zhou, Ouyang, Wang \&
  Zhang}]{za}
\bibinfo{author}{Zhang, R.}, \bibinfo{author}{Zhou, J.},
  \bibinfo{author}{Ouyang, S.}, \bibinfo{author}{Wang, X.}, \&
  \bibinfo{author}{Zhang, H.} (\bibinfo{year}{2013}{\natexlab{b}}).
\newblock \bibinfo{title}{{ Optimal operation of multireservoir system by
  multi-elite guide particle swarm optimization}}.
\newblock {\it \bibinfo{journal}{International Journal of Electrical Power \&
  Energy Systems}\/},  {\it \bibinfo{volume}{48}\/}, \bibinfo{pages}{58--68}.
  \DOIprefix\doi{https://doi.org/10.1016/j.ijepes.2012.11.031}.
\bibitem[{Zhang et~al.(2018)Zhang, Le, Liao, Zheng, Liu \& An}]{hu}
\bibinfo{author}{Zhang, Y.}, \bibinfo{author}{Le, J.}, \bibinfo{author}{Liao,
  X.}, \bibinfo{author}{Zheng, F.}, \bibinfo{author}{Liu, K.}, \&
  \bibinfo{author}{An, X.} (\bibinfo{year}{2018}).
\newblock \bibinfo{title}{{ Multi-objective hydro-thermal-wind coordination
  scheduling integrated with large-scale electric vehicles using IMOPSO}}.
\newblock {\it \bibinfo{journal}{Renewable Energy}\/},  {\it
  \bibinfo{volume}{128}\/}, \bibinfo{pages}{1--19}.
  \DOIprefix\doi{https://doi.org/10.1016/j.renene.2018.05.067}.
\bibitem[{Zhang et~al.(2020)Zhang, Wang, Li, Yeh, Jian \& Dong}]{nr11}
\bibinfo{author}{Zhang, Y.}, \bibinfo{author}{Wang, G.-G.},
  \bibinfo{author}{Li, K.}, \bibinfo{author}{Yeh, W.-C.},
  \bibinfo{author}{Jian, M.}, \& \bibinfo{author}{Dong, J.}
  (\bibinfo{year}{2020}).
\newblock \bibinfo{title}{Enhancing moea/d with information feedback models for
  large-scale many-objective optimization}.
\newblock {\it \bibinfo{journal}{Information Sciences}\/},  {\it
  \bibinfo{volume}{522}\/}, \bibinfo{pages}{1--16}.
  \DOIprefix\doi{doi.org/10.1016/j.ins.2020.02.066}.
\bibitem[{Zheng et~al.(2001)Zheng, Simpson \& Zecchin}]{zeng}
\bibinfo{author}{Zheng, F.}, \bibinfo{author}{Simpson, A.}, \&
  \bibinfo{author}{Zecchin, A.} (\bibinfo{year}{2001}).
\newblock \bibinfo{title}{{A combined NLP-differential evolution algorithm
  approach for the optimization of looped water distribution systems}}.
\newblock {\it \bibinfo{journal}{Water Resources Research}\/},  {\it
  \bibinfo{volume}{47}\/}, \bibinfo{pages}{1--18}.
  \DOIprefix\doi{https://doi.org/10.1029/2011WR010394}.
\bibitem[{Zhou et~al.(2011)Zhou, Qu, Li, Zhao, Suganthan \&
  Zhang}]{zhou2011multiobjective}
\bibinfo{author}{Zhou, A.}, \bibinfo{author}{Qu, B.-Y.}, \bibinfo{author}{Li,
  H.}, \bibinfo{author}{Zhao, S.-Z.}, \bibinfo{author}{Suganthan, P.~N.}, \&
  \bibinfo{author}{Zhang, Q.} (\bibinfo{year}{2011}).
\newblock \bibinfo{title}{Multiobjective evolutionary algorithms: A survey of
  the state of the art}.
\newblock {\it \bibinfo{journal}{Swarm and Evolutionary Computation}\/},  {\it
  \bibinfo{volume}{1}\/}, \bibinfo{pages}{32--49}.
\bibitem[{Zitzler et~al.(2001)Zitzler, Laumanns \& Thiele}]{spea2}
\bibinfo{author}{Zitzler, E.}, \bibinfo{author}{Laumanns, L.}, \&
  \bibinfo{author}{Thiele, M.} (\bibinfo{year}{2001}).
\newblock \bibinfo{title}{{SPEA2: Improving the strength pareto evolutionary
  algorithm}}.
\newblock {\it \bibinfo{journal}{TIK-Report 103, May}\/}, .

\end{thebibliography}

\end{document}